\PassOptionsToPackage{unicode}{hyperref}
\PassOptionsToPackage{hyphens}{url}
\PassOptionsToPackage{dvipsnames,svgnames,x11names}{xcolor}
\documentclass[
  11pt,
]{article}
\usepackage{xcolor}
\usepackage[a4paper,margin=2.2cm]{geometry}
\usepackage{amsmath,amssymb}
\usepackage{iftex}
\ifPDFTeX
  \usepackage[T1]{fontenc}
  \usepackage[utf8]{inputenc}
  \usepackage{textcomp} % provide euro and other symbols
\else % if luatex or xetex
  \usepackage{unicode-math} % this also loads fontspec
  \defaultfontfeatures{Scale=MatchLowercase}
  \defaultfontfeatures[\rmfamily]{Ligatures=TeX,Scale=1}
\fi
\usepackage{lmodern}
\ifPDFTeX\else
\fi
\IfFileExists{upquote.sty}{\usepackage{upquote}}{}
\IfFileExists{microtype.sty}{% use microtype if available
  \usepackage[]{microtype}
  \UseMicrotypeSet[protrusion]{basicmath} % disable protrusion for tt fonts
}{}
\makeatletter
\@ifundefined{KOMAClassName}{% if non-KOMA class
  \IfFileExists{parskip.sty}{%
    \usepackage{parskip}
  }{% else
    \setlength{\parindent}{0pt}
    \setlength{\parskip}{6pt plus 2pt minus 1pt}}
}{% if KOMA class
  \KOMAoptions{parskip=half}}
\makeatother
\usepackage{color}
\usepackage{fancyvrb}

\DefineVerbatimEnvironment{Highlighting}{Verbatim}{commandchars=\\\{\}}
\newenvironment{Shaded}{}{}

\newcommand{\AttributeTok}[1]{\textcolor[rgb]{0.49,0.56,0.16}{#1}}

\newcommand{\CommentTok}[1]{\textcolor[rgb]{0.38,0.63,0.69}{\textit{#1}}}

\newcommand{\ExtensionTok}[1]{#1}

\newcommand{\NormalTok}[1]{#1}

\newcommand{\StringTok}[1]{\textcolor[rgb]{0.25,0.44,0.63}{#1}}

\usepackage{longtable,booktabs,array}
\usepackage{calc} % for calculating minipage widths
\usepackage{etoolbox}
\makeatletter
\patchcmd\longtable{\par}{\if@noskipsec\mbox{}\fi\par}{}{}
\makeatother
\IfFileExists{footnotehyper.sty}{\usepackage{footnotehyper}}{\usepackage{footnote}}
\makesavenoteenv{longtable}
\usepackage{graphicx}
\makeatletter
\newsavebox\pandoc@box
\newcommand*\pandocbounded[1]{% scales image to fit in text height/width
  \sbox\pandoc@box{#1}%
  \Gscale@div\@tempa{\textheight}{\dimexpr\ht\pandoc@box+\dp\pandoc@box\relax}%
  \Gscale@div\@tempb{\linewidth}{\wd\pandoc@box}%
  \ifdim\@tempb\p@<\@tempa\p@\let\@tempa\@tempb\fi% select the smaller of both
  \ifdim\@tempa\p@<\p@\scalebox{\@tempa}{\usebox\pandoc@box}%
  \else\usebox{\pandoc@box}%
  \fi%
}
\def\fps@figure{htbp}
\makeatother
\providecommand{\tightlist}{%
  \setlength{\itemsep}{0pt}\setlength{\parskip}{0pt}}
\usepackage[]{natbib}
\usepackage{amssymb}   % \checkmark and friends
\usepackage{pifont}    % \ding{51} / \ding{55} — the ✓ / ✗ pair (psnfss, stock)

\DeclareUnicodeCharacter{2713}{\ding{51}}              % ✓ check mark
\DeclareUnicodeCharacter{2717}{\ding{55}}              % ✗ ballot X
\DeclareUnicodeCharacter{2264}{\ensuremath{\leq}}      % ≤
\DeclareUnicodeCharacter{2265}{\ensuremath{\geq}}      % ≥
\DeclareUnicodeCharacter{2212}{\ensuremath{-}}         % − minus sign
\DeclareUnicodeCharacter{0394}{\ensuremath{\Delta}}    % Δ
\DeclareUnicodeCharacter{03B5}{\ensuremath{\varepsilon}} % ε
\DeclareUnicodeCharacter{03B1}{\ensuremath{\alpha}}    % α
\usepackage{bookmark}
\IfFileExists{xurl.sty}{\usepackage{xurl}}{} % add URL line breaks if available
\hypersetup{
  pdftitle={Confidently Wrong: Exception Chain Collapse in Frontier LLM Rule Evaluation},
  pdfauthor={Paul Simpson; John Kozak; Lisa Doake},
  colorlinks=true,
  linkcolor={black},
  filecolor={Maroon},
  citecolor={Blue},
  urlcolor={blue},
  pdfcreator={LaTeX via pandoc}}

\title{Confidently Wrong: Exception Chain Collapse in Frontier LLM Rule
Evaluation}
\usepackage{etoolbox}
\makeatletter
\providecommand{\subtitle}[1]{% add subtitle to \maketitle
  \apptocmd{\@title}{\par {\large #1 \par}}{}{}
}
\makeatother
\subtitle{A Neuro-Symbolic Architecture for Deterministic Eligibility
Determination}
\author{Paul Simpson \and John Kozak \and Lisa Doake}
\date{Aethis (aethis.ai) · July 2026}

\begin{document}
\maketitle
\begin{abstract}
We document a failure class in frontier large language models ---
exception chain collapse --- observed during eligibility evaluation
under nested conditional rules of the form ``A is required UNLESS B
applies, UNLESS C overrides B''. The failure is real and reproducible at
first observation, but its empirical surface is unstable: between March
and April 2026, several reported failure cells closed silently under the
same model alias, with no version bump (GPT-5.4 on construction
insurance moved from 96.6\% to 100\% on the same prompt and harness).
For a regulated workflow that depends on benchmark-time accuracy claims,
this is the central problem: frontier-model accuracy is a moving
compliance boundary, and it moves without notice.

We present the Aethis Eligibility Module, a neuro-symbolic architecture
that uses LLMs to author rules from authoritative sources, then executes
those rules through an SMT-based deterministic evaluation layer. The
execution layer is exactly consistent with the authored specification,
independent of model drift, reasoning-effort defaults, or prompt format.
Three evidence bases support the claim. (i) A controlled benchmark of
225 scenarios across four regulatory domains documents the original
failure pattern and, in replication, the silent drift that partially
closed it (§6). (ii) A 20-scenario adversarial extension on construction
insurance (§6.4.1): the deterministic engine scores 20/20; of four
frontier-LLM configurations, one (GPT-5.4 at low reasoning effort) also
scores 20/20, while the other three --- including Anthropic's strongest
model at evaluation time --- each fail the same coverage-gap edge case.
Matching the engine on a given slice is achievable; staying matched
across silent model updates and configuration changes is the property
none of these systems can offer. (iii) External validation on nine
peer-reviewed LegalBench tasks, 949 held-out cases (§6.10): the engine
is significantly more accurate than each of three frontier models
(combined McNemar's \(p \le 0.003\)), with the largest margins against
the Anthropic models --- up to +41 percentage points --- on the curated
multi-prong rule-application tasks (GPT-5.4's larger margins are
dominated by a prompt-format sensitivity analysed in §6.10.4).

The architectural contribution is to relocate uncertainty from the
inference boundary, where it is silent and continuous, to the
specification boundary, where it is deliberate and audited (§5.5, §9.4).
All numbers are reproducible from committed scripts and result JSONs;
the scenarios, rule encodings, and harness are public.
\end{abstract}

{
\setcounter{tocdepth}{2}
\tableofcontents
}
\begin{center}\rule{0.5\linewidth}{0.5pt}\end{center}

\section{1. Introduction}\label{introduction}

In 1986, Sergot, Sadri, Kowalski, and colleagues published a landmark
paper in \emph{Communications of the ACM}: they had formalised the
British Nationality Act 1981 as a Prolog logic program
\citeyearpar{sergot1986bna}. The paper demonstrated that statutory
legislation - with its characteristic structure of routes, exemptions,
and override provisions - could be faithfully encoded as Horn clauses
with negation as failure, and evaluated with provable correctness. It
was a foundational result in computational law: legislation is logic,
and logic can be executed.

Forty years on, large language models can read the same Act, explain it
in plain English, and answer broad legal questions with fluency that
would have seemed extraordinary even five years ago. It is tempting to
conclude that the formal encoding project is no longer necessary. Our
benchmark tests this assumption directly on the same legislation Sergot
et al.~encoded, extended to three further domains, and finds it does not
hold - at least not for the specific class of task where exception
chains are nested three levels deep.

The failure is specific. On straightforward multi-route eligibility
logic, frontier models perform well, achieving 100\% accuracy on 43
English-language scenarios (95\% Wilson CI {[}91.8\%, 100\%{]}). But
when legislation introduces nested exception chains, accuracy degrades
sharply and in a way that is insensitive to temperature, sample count,
and the prompting strategies tested here (a single-prompt baseline and
one enhanced variant; see §6.7). In the March 2026 baseline, Claude Opus
4.6 scored 61/68 (89.7\%, CI {[}80.2\%, 94.9\%{]}) on the spacecraft
section. With 10 runs on each of the 7 then-failing scenarios --- 70
trials in total --- it produced zero correct answers. The
Clopper--Pearson 95\% one-sided upper bound on per-trial success
probability is 4.19\%: at that snapshot the failures were systematic,
not stochastic. (The April 2026 replication finds most of those specific
cells have since closed under the same model alias --- the instability
analysed in §6.5 Finding 6 and the §6.7 caveats.) Attempts to close the
gap via enhanced prompting trade false negatives for false positives:
the enhanced prompt reduces net accuracy to 64.7\% (CI {[}52.8\%,
75.0\%{]}) while introducing 20 false positives for the first time.

More precisely, the failure regime characterised here is
\emph{multi-prong compositional rule evaluation under nested
exceptions}: tasks where multiple coordinate clauses must each be
evaluated against a typed input and then combined across two or more
levels of UNLESS-style exception structure. The ``exception chain
collapse'' terminology is retained throughout the paper as the dominant
failure pattern within that regime; the more precise scope phrase is
what we mean wherever the dominant pattern is invoked.

We make three contributions:

\begin{enumerate}
\def\labelenumi{\arabic{enumi}.}
\item
  \textbf{A failure pattern taxonomy.} We identify and characterise two
  systematic failure patterns in nested exception-chain evaluation:
  \emph{exemption anchoring} (failure to evaluate alternative routes
  independently when the primary route fails) and \emph{exception chain
  collapse} (failure to correctly nest multi-level UNLESS logic). We
  observe both patterns across eight models (four frontier, four
  production-tier) and two providers on this benchmarked class of task,
  on the model--suite combinations actually evaluated (§6.1 coverage
  matrix); individual models pass individual suites, and specific
  failure cells move across model updates (§6.5 Findings 4 and 6).
\item
  \textbf{A multi-domain benchmark.} We present 225 scenarios across
  four sections spanning two UK immigration requirements (life-in-the-UK
  knowledge and English language proficiency), a synthetic spacecraft
  certification statute, and a synthetic construction insurance wording
  modelled on London market DE3/DE5 clause structure - designed to
  isolate exception chain evaluation as the test variable. Benchmark
  scenarios are released as a public dataset. The specific accuracy
  figures reported here will evolve as models improve; the task
  structure and failure pattern constitute the durable contribution.
\item
  \textbf{The Aethis Eligibility Module.} We describe a neuro-symbolic
  architecture that uses LLMs for rule authoring and an SMT-based
  constraint evaluation engine for rule execution, achieving complete
  consistency with the benchmark's formal rule fixtures across all 225
  scenarios, with \textless1ms evaluation latency and near-zero marginal
  cost after compilation.
\end{enumerate}

The paper is structured as follows. Section 2 summarises key findings.
Section 3 reviews related work. Section 4 characterises the failure
pattern. Section 5 describes the architecture. Section 6 presents the
benchmark. Section 7 discusses challenges in LLM-guided rule synthesis.
Sections 8, 9, and 10 address generalisation, compliance, and
limitations.

\begin{center}\rule{0.5\linewidth}{0.5pt}\end{center}

\section{2. Summary of Contributions}\label{summary-of-contributions}

Artificial intelligence is entering high-stakes decision-making:
immigration eligibility, safety certification, insurance underwriting,
benefits entitlement, financial compliance. These are domains where
errors have material consequences, where false negatives deny people
their rights, and where explainability is mandatory.

We present the \textbf{Aethis Eligibility Module} (hereafter the
Eligibility Module; where unambiguous, simply the Module or the engine),
a neuro-symbolic engine that separates what LLMs do well from what
requires formal guarantees. LLMs read authoritative sources and generate
rules as structured code. The Eligibility Module compiles and evaluates
those rules using an SMT-based deterministic evaluation layer, providing
constraint evaluation with mathematically defined semantics and full
auditability.

The architectural contribution is to relocate the locus of uncertainty:
from the inference boundary, where it is silent and continuous, to the
specification boundary, where it is deliberate and audited. This does
not eliminate the underlying uncertainty; it makes it tractable ---
addressable through SME-validated test suites, explicit bundle
versioning, and the L1/L2/L3 separation set out in Section 5.5.

Our benchmark of 225 scenarios across four sections tests eight LLMs
(four frontier, four production-tier) against the Eligibility Module on
the specific task of nested exception-chain evaluation --- six models in
the March 2026 pass, with Claude Opus 4.7 and GPT-4.1-mini added in the
April 2026 v3.8 replication. The Module achieves complete consistency
with the benchmark's formal rule fixtures across all domains - a
consequence of deterministic execution over the authored specification,
not empirical tuning. On adversarial exception-chain scenarios, frontier
models produce systematic false negatives: in the March 2026 baseline
Claude Opus 4.6 returned the wrong answer on 10\% of spacecraft
scenarios (1.5\% by the April 2026 replication --- the drift §6
documents), and in the March 2026 baseline evaluations all observed
failures are false negatives - eligible applicants incorrectly rejected,
valid claims incorrectly denied (later evaluation arms do surface false
positives; see §6.5 Finding 2). Attempting to improve LLM accuracy via
enhanced prompting trades false negatives for false positives: the
enhanced prompt reduces false negatives from 7 to 4 but introduces 20
false positives, reducing net accuracy from 90\% to 65\%.

On the benchmarked class of nested exception-chain tasks, LLM errors are
systematic enough that deterministic formal execution removes one entire
class of error from the trust surface for high-stakes decisions where
consistency with formal fixtures is required. This is not a claim that
LLMs are unreliable in general; it is a specific finding about a
specific class of rule evaluation.\footnote{Aethis (aethis.ai) is
  deploying the system described in this paper in a controlled UK
  immigration pilot: it prepares eligibility evaluations for solicitor
  review, and solicitors remain the decision-makers. The immigration
  benchmark sections cover selected requirements from this domain; the
  full determination involves additional sections not included in this
  publication.}

\begin{center}\rule{0.5\linewidth}{0.5pt}\end{center}

\section{3. Related Work}\label{related-work}

This work sits at the intersection of four research traditions: formal
and computational approaches to legal reasoning; empirical evaluation of
LLM logical reasoning; the limits of prompt engineering as a reliability
strategy; and neuro-symbolic architectures that combine LLM fluency with
formal execution guarantees.

\subsection{3.1 Formal and Computational Approaches to Legal
Reasoning}\label{formal-and-computational-approaches-to-legal-reasoning}

The challenge of encoding legislation as executable logic has a
forty-year history. Sergot et al. \citeyearpar{sergot1986bna} formalised
the British Nationality Act 1981 as a Prolog logic program,
demonstrating that statutory rules could be faithfully represented as
Horn clauses with negation as failure. Published in \emph{Communications
of the ACM} in 1986, this work identified the same legislation evaluated
in our benchmark and showed that OR-branching eligibility logic can be
captured in a formal system with provable properties. The Module
revisits the same statutory text with a different technical foundation -
formal constraints compiled from LLM-authored rules rather than
hand-coded Prolog - and extends evaluation to adversarial exception
chain scenarios not part of the original formalism.

McCarty's TAXMAN system \citeyearpar{mccarty1977taxman} demonstrated as
early as 1977 that AI systems could reason over tax code with explicit
logical representations. Bench-Capon and colleagues developed
value-based argumentation frameworks for legal reasoning over multiple
decades \citeyearpar{benchcapon2010argument}, establishing that
legislation's exception structure requires more than propositional logic
to represent faithfully.

The formal treatment for exception chains specifically is defeasible
logic, introduced by Reiter \citeyearpar{reiter1980default} and
developed by Nute \citeyearpar{nute1994defeasible} and Governatori et
al. \citeyearpar{governatori2010changing}. Defeasible logic provides
formal semantics for ``A holds UNLESS B applies, UNLESS C overrides B''
- precisely the pattern our benchmark identifies as a failure pattern
for LLMs. The Module does not use defeasible logic directly, instead
compiling exception chains to formal material implication constraints
evaluated by an SMT solver, but operates in the same tradition: the
failure pattern we document is exactly the problem defeasible logic was
designed to solve, now re-emerging in systems that replaced formal
encoding with statistical inference over legislative text.

The broader ``Rules as Code'' movement --- encoding legislation as
machine-executable representations at the point of drafting --- has
gained institutional momentum since 2020, with government-backed
programmes in New Zealand, France, and Germany deploying visual
decision-tree modelling tools for legislative analysis
\citep{mohun2020cracking, mowbray2023representing}. These systems encode
statutory logic as structured flowcharts or decision trees authored via
no-code configuration, enabling non-technical legislative drafters to
model requirements directly. Mowbray et al.
\citeyearpar{mowbray2023representing} identify scalability as the
central challenge: decision trees grow combinatorially as exception
depth increases, and maintaining consistency across large rule sets
requires manual audit. The approach shares this paper's core premise ---
that statutory rules should be formally encoded rather than
statistically inferred --- but differs in representational power.
Decision trees can model branching logic; constraint-based
representations can additionally prove properties of the encoded rule
system (completeness, consistency, unreachability of dead states). The
exception-chain collapse pattern documented in §4 is precisely the class
of nested structure where this distinction is practically relevant: a
three-level exception chain is representable as a decision tree, but
verifying that no branch combination produces a contradictory outcome
requires the kind of satisfiability check that decision-tree
representations do not natively support.

\subsection{3.2 LLM Benchmarks for Legal and Logical
Reasoning}\label{llm-benchmarks-for-legal-and-logical-reasoning}

LegalBench \citeyearpar{guha2023legalbench} provides the most
comprehensive evaluation of LLM performance on legal tasks, testing 162
tasks across six categories. We report external validation on a subset
of LegalBench tasks in §6.10. The two benchmarks remain complementary in
scope: LegalBench tests breadth of legal reasoning across 162 tasks; the
benchmark in §6.1--§6.9 isolates a single failure pattern under
controlled conditions. §6.10 evaluates the architecture on nine
LegalBench tasks selected to span the failure pattern (multi-prong rule
application) and unbiased-sample tasks (single-clause classification).

FOLIO \citeyearpar{han2022folio} tests natural language inference
grounded in first-order logic, demonstrating that LLMs struggle with
tasks requiring explicit formal logical structure even when relevant
premises are provided. Our findings are consistent: the failures we
observe are not information retrieval failures (the legislation is
provided in full) but reasoning failures arising from the compositional
structure of the task.

\subsection{3.3 Systematic Limits of LLM
Reasoning}\label{systematic-limits-of-llm-reasoning}

Dziri et al. \citeyearpar{dziri2023faith} demonstrate fundamental limits
of transformer architectures on compositional tasks, showing that
performance degrades systematically as task depth increases even when
each individual step is within the model's capability. Exception chain
evaluation is compositional in precisely this sense: each individual
rule is simple, but the nested application of three independent
exception conditions requires compositional reasoning over the rule
structure. Our finding that failures concentrate on three-level
exception chains and are absent on two-level OR-branching is consistent
with the compositional depth hypothesis.

Shi et al. \citeyearpar{shi2023distracted} show that LLMs are
systematically distracted by irrelevant context. The related phenomenon
of \emph{exemption anchoring} - where attention on the failed primary
route suppresses evaluation of valid alternative routes - may be a
manifestation of the same attentional bias in a rule-following context.

Valmeekam et al. \citeyearpar{valmeekam2022planning} demonstrate that
LLMs fail reliably on planning tasks requiring state tracking and
logical consistency, not through random error but through systematic
misapplication of learned heuristics. Our finding that Claude Opus 4.6
produces 0/10 correct responses across ten independent runs on veteran
exemption scenarios reflects the same pattern: consistent, confident,
and wrong.

Valmeekam, Stechly and Kambhampati \citeyearpar{valmeekam2024lrms}
extend this work to reasoning-optimised language models, reporting that
OpenAI's o1 degrades sharply on harder Mystery Blocksworld variants and
obfuscated planning tasks --- the failure mode is not eliminated by
inference-time reasoning compute, merely shifted to deeper compositional
depth. We initially read this as parallel to a v3.6 / v3.7 intra-model
reasoning-effort finding on the construction-CAR exception chain; that
finding was withdrawn in v3.8 (see §6.5 Finding 5) after the original
result failed to replicate under instrumented testing. The Valmeekam et
al.~result remains relevant context for §6.9's pre-registered N=66
reasoning-effort replication.

Mirzadeh et al. \citeyearpar{mirzadeh2024gsmsymbolic} show that LLM
mathematical reasoning is brittle to surface-form changes: renaming
entities or adding irrelevant clauses to GSM8K problems systematically
degrades accuracy, even on frontier models. This is evidence that what
looks like symbolic reasoning is substantially pattern matching over
learned surface forms. The exception-chain failure pattern we observe is
in the same family --- LLMs produce fluent legal-sounding justifications
that do not track the actual logical structure of the rule.

\subsection{3.4 Prompt Engineering and Its
Limits}\label{prompt-engineering-and-its-limits}

Chain-of-thought prompting \citeyearpar{wei2022cot} and zero-shot
reasoning elicitation \citeyearpar{kojima2022zeroshot} have
substantially improved LLM performance on arithmetic and multi-step
problems. Wang et al. \citeyearpar{wang2023selfconsistency} show that
self-consistency (majority voting) improves performance on reasoning
tasks. Our robustness analysis (Section 6.7) explicitly tests whether
enhanced prompting closes the accuracy gap on exception chain
evaluation, and finds a trade-off rather than a fix: an enhanced prompt
that correctly identifies and targets the failure pattern (independent
exemption evaluation) reduces false negatives from 7 to 4 while
introducing 20 false positives, reducing net accuracy from 90\% to 65\%.
The result demonstrates that prompt-based repair on this class of task
is fragile - instructions that correct under-application of exemptions
simultaneously cause over-application elsewhere.

\subsection{3.5 Neuro-Symbolic Architectures and LLM + Formal Method
Hybrids}\label{neuro-symbolic-architectures-and-llm-formal-method-hybrids}

The neuro-symbolic research programme \citeyearpar{garcez2009neural}
argues that robust AI systems require integration of neural pattern
recognition with symbolic reasoning. Marcus
\citeyearpar{marcus2020nextdecade} argues that the reliability
limitations of purely statistical systems necessitate a return to hybrid
approaches combining learned representations with structured reasoning.
Kambhampati et al. \citeyearpar{kambhampati2024llmmodulo} advance this
position with the \emph{LLM-Modulo} framework, arguing that LLMs are
most robustly deployed as approximate generators paired with formal
verifiers and critics that provide external correctness guarantees. The
Module is a specific instantiation of the LLM-Modulo pattern applied to
statutory rule evaluation: LLMs perform pattern-recognition tasks they
excel at (reading legislation, extracting structure, generating code),
while a constraint evaluation engine handles the evaluation task
requiring mathematical guarantees.

Statutory rule evaluation is a particularly favourable application of
the LLM-Modulo pattern, in ways that distinguish it from prior
instantiations in code synthesis and symbolic planning. Three properties
combine. First, the verification fragment is decidable: rule application
over compiled constraints terminates deterministically with a total
correctness criterion, in contrast to test-suite verification of
synthesised code (which provides only partial coverage) or constraint
satisfaction over learned world models (which is frequently
approximate). Second, the artefact being verified --- a compiled rule
bundle --- is persistent and amortised: a single authoring pass serves
arbitrarily many subsequent decisions, where per-query LLM-Modulo cycles
in code synthesis or planning re-incur authoring cost on every instance.
Third, the correctness criterion is externally specified by statute and
domain-expert review rather than chosen by the system designer, which
makes test-driven validation against expert-defined fixtures (Section
7.3) a meaningful integrity check rather than a tautology. These
properties together explain why the LLM-Modulo separation can deliver
categorical guarantees at the execution layer in this domain, even where
the same separation provides only best-effort guarantees in general
program synthesis or planning.

Most directly related is Logic-LM \citeyearpar{pan2023logiclm}, which
uses LLMs to translate natural language problems into formal logical
representations, then invokes symbolic solvers for evaluation. LINC
\citeyearpar{olausson2023linc} similarly uses LLMs to generate
first-order logic programs from natural language for theorem prover
evaluation. These systems demonstrate the feasibility of the
authoring-execution separation that underlies the Eligibility Module.
The present system differs in three respects relevant to high-stakes
deployment: it operates on \emph{persistently stored} rule bundles
rather than ephemeral per-query translations; it maintains a provenance
chain linking each rule to specific source citations; and it is designed
for production deployment where audit trails and version control are
compliance requirements.

Program-aided language models (PAL \citeyearpar{gao2023pal}) demonstrate
the broader pattern of using LLMs to generate code that is then executed
deterministically. The Module applies this separation to statutory rule
encoding with additional quality engineering (Section 7) to ensure
generated rules meet a quality threshold before entering the persistent
rule store.

\subsection{3.6 SMT-Based Constraint
Evaluation}\label{smt-based-constraint-evaluation}

Satisfiability Modulo Theories (SMT) solving has seen extensive
application in software verification, symbolic execution, hardware
design, and safety-critical systems. The Module applies SMT-based
constraint evaluation to regulatory rule execution - rules compiled from
LLM-authored code into formal constraint representations. The near-zero
marginal evaluation cost after compilation makes formal constraint
evaluation practical for production deployment at scale, a property that
distinguishes constraint compilation from per-query LLM inference.

\begin{center}\rule{0.5\linewidth}{0.5pt}\end{center}

\section{4. The Problem: LLMs and Nested Exception-Chain
Evaluation}\label{the-problem-llms-and-nested-exception-chain-evaluation}

\subsection{4.1 What Makes a Decision
``High-Stakes''}\label{what-makes-a-decision-high-stakes}

High-stakes decisions share three properties that distinguish them from
general AI tasks:

\begin{enumerate}
\def\labelenumi{\arabic{enumi}.}
\tightlist
\item
  \textbf{Material consequence.} An incorrect determination can deny
  someone a right (citizenship, a benefit, a licence), expose an
  organisation to regulatory action, or cause financial harm.
\item
  \textbf{Auditability requirement.} A regulator, court, or oversight
  body may require an explanation of how the determination was reached,
  traceable to specific rules or statutory provisions.
\item
  \textbf{OR-branching logic.} Rules typically provide multiple
  independent pathways to satisfaction (primary routes, exemptions,
  waivers, overrides), each of which is independently sufficient.
\end{enumerate}

Immigration law illustrates this structure. The British Nationality Act
1981 sets out multiple requirements for naturalisation, some of which
include alternative pathways to satisfaction --- such as age-based
exemptions, medical exemptions, and discretionary waivers --- which
operate as independent routes within those requirements (disjunctive
branches), not as standalone routes to overall eligibility (which
requires conjunction across requirements).

\subsection{4.2 Two Failure Patterns}\label{two-failure-patterns}

When a large language model is asked to determine eligibility, it
processes legislation and applicant data as a single reasoning task. Our
benchmark reveals two systematic failure patterns on nested
exception-chain tasks in this benchmarked setting. We make no claim that
these patterns generalise to all legal reasoning; they are specific to
the class of task where exception chains are nested three levels deep:

\textbf{Failure Pattern 1: Exemption anchoring.} LLMs treat exemption
and waiver routes as secondary to the primary pathway rather than as
independently sufficient alternatives. When the primary route fails, the
LLM anchors on that failure and discounts exemptions.

\textbf{Failure Pattern 2: Exception chain collapse.} When rules contain
multi-level exception chains (``A is required UNLESS B applies, UNLESS C
overrides B''), LLMs fail to correctly evaluate the nested logic.

\textbf{Table 1: Multi-Level Exception Chain (Spacecraft Benchmark)}

The Spacecraft Crew Certification Act (a synthetic statute modelled on
UK legislative structure) contains a three-level exception chain for
flight readiness:

{\def\LTcaptype{none} % do not increment counter
\begin{longtable}[]{@{}
  >{\centering\arraybackslash}p{(\linewidth - 4\tabcolsep) * \real{0.2593}}
  >{\raggedright\arraybackslash}p{(\linewidth - 4\tabcolsep) * \real{0.2222}}
  >{\raggedright\arraybackslash}p{(\linewidth - 4\tabcolsep) * \real{0.5185}}@{}}
\toprule\noalign{}
\begin{minipage}[b]{\linewidth}\centering
Level
\end{minipage} & \begin{minipage}[b]{\linewidth}\raggedright
Rule
\end{minipage} & \begin{minipage}[b]{\linewidth}\raggedright
Plain English
\end{minipage} \\
\midrule\noalign{}
\endhead
\bottomrule\noalign{}
\endlastfoot
\textbf{Base} & Flight readiness required (500hrs + licence) & Must have
500+ hours AND a pilot licence \\
\textbf{Exception A} & Age \textgreater= 60 exempts from flight
readiness & Over-60s don't need flight readiness\ldots{} \\
\textbf{Exception B} & \ldots UNLESS mission is orbital & \ldots but
orbital missions revoke the age exemption \\
\textbf{Override C} & 1000+ flight hours overrides everything & Veteran
pilots (1000+ hrs) are always exempt \\
\end{longtable}
}

\textbf{Table 2: Failures on Exception Chain Scenarios (spacecraft,
adversarial suite)}

{\def\LTcaptype{none} % do not increment counter
\begin{longtable}[]{@{}
  >{\raggedright\arraybackslash}p{(\linewidth - 10\tabcolsep) * \real{0.5312}}
  >{\centering\arraybackslash}p{(\linewidth - 10\tabcolsep) * \real{0.0938}}
  >{\centering\arraybackslash}p{(\linewidth - 10\tabcolsep) * \real{0.0938}}
  >{\centering\arraybackslash}p{(\linewidth - 10\tabcolsep) * \real{0.0938}}
  >{\centering\arraybackslash}p{(\linewidth - 10\tabcolsep) * \real{0.0938}}
  >{\centering\arraybackslash}p{(\linewidth - 10\tabcolsep) * \real{0.0938}}@{}}
\toprule\noalign{}
\begin{minipage}[b]{\linewidth}\raggedright
Scenario
\end{minipage} & \begin{minipage}[b]{\linewidth}\centering
Expected
\end{minipage} & \begin{minipage}[b]{\linewidth}\centering
Opus 4.6
\end{minipage} & \begin{minipage}[b]{\linewidth}\centering
Sonnet 4.6
\end{minipage} & \begin{minipage}[b]{\linewidth}\centering
GPT-5.4
\end{minipage} & \begin{minipage}[b]{\linewidth}\centering
GPT-5-mini
\end{minipage} \\
\midrule\noalign{}
\endhead
\bottomrule\noalign{}
\endlastfoot
Age 25, 1500hrs, no licence, suborbital & Eligible & 1/3 & 0/3 & 3/3 &
1/3 \\
Age 59, 1000hrs, no licence & Eligible & 0/3 & 0/3 & 3/3 & 1/3 \\
Age 60, orbital, 999hrs, WITH licence & Eligible & 0/3 & 3/3 & 3/3 &
1/3 \\
Age 22, 1001hrs, no licence, orbital + rad cert & Eligible & 0/3 & 3/3 &
3/3 & 1/3 \\
Dolphin*, 1200hrs veteran, orbital, provider medical & Eligible & 0/3 &
3/3 & 3/3 & 1/3 \\
\end{longtable}
}

\emph{*Dolphin is a valid species under the synthetic statute (s.3
excludes only Vogons). This scenario tests whether the model correctly
applies the veteran exemption to a non-human applicant with otherwise
valid credentials.}

The veteran exemption (Override C) is the hardest concept for LLMs in
this benchmark. It operates independently of age: a 25-year-old with
1500 flight hours is exempt from flight readiness requirements, just as
a 60-year-old would be. Most LLMs tested treat the veteran exemption as
age-dependent, producing false negatives with high confidence --- Claude
Opus 4.6 and Sonnet 4.6 mark the age-59 veteran ineligible on every run
(0/3; Table 2), though GPT-5.4 answers the veteran scenarios correctly.
The Eligibility Module evaluates these correctly (§5.4).

\subsection{4.3 Why False Negatives
Matter}\label{why-false-negatives-matter}

In high-stakes decision-making, a false negative means wrongly telling
someone they do not qualify. Exemptions exist specifically for edge
cases - the people who cannot satisfy the standard route but qualify
through an alternative path. These are precisely the applicants most
likely to be wrongly denied. On the benchmarked exception-chain
scenarios, the failures are systematic: Claude Opus 4.6 returns
``ineligible'' on every run (0/3) for the age-59 veteran scenario, and
on all runs of most other veteran-exemption scenarios in Table 2.
Majority voting would not catch this.

Our construction insurance benchmark (Section 6.4) demonstrates the same
failure pattern in a different domain: a CAR policy defect exclusion
clause with a five-level exception chain. GPT-5.4, which achieves 100\%
on the spacecraft scenarios, dropped to 96.6\% on the insurance
scenarios in the March 2026 snapshot (a cell that subsequently closed
under model drift; §6.5 Finding 4). The pattern is not specific to a
single model, provider, or domain.

\begin{center}\rule{0.5\linewidth}{0.5pt}\end{center}

\section{5. The Neuro-Symbolic
Architecture}\label{the-neuro-symbolic-architecture}

\subsection{5.1 Design Principle}\label{design-principle}

The architecture is built on a simple observation: LLMs and formal
constraint evaluation excel at different tasks. LLMs are exceptional at
reading legislation, understanding context, and generating structured
representations. SMT constraint solvers are exceptional at evaluating
logical expressions with mathematical guarantees.

The \textbf{Aethis Eligibility Module} separates these concerns:

\begin{itemize}
\tightlist
\item
  \textbf{Phase 1: Rule Authoring (LLM).} Read authoritative legal
  sources, discover sections and fields, generate rules as structured
  code, evaluate quality iteratively.
\item
  \textbf{Phase 2: Rule Execution (Eligibility Module).} Parse generated
  code, compile to formal SMT constraints, evaluate against applicant
  data with deterministic execution.
\end{itemize}

\subsection{5.2 Phase 1: Automated Rule
Authoring}\label{phase-1-automated-rule-authoring}

The rule authoring pipeline transforms authoritative legal text into
executable compliance rules:

\textbf{Figure 1: End-to-End Pipeline}

\pandocbounded{\includegraphics[keepaspectratio,alt={End-to-End Pipeline}]{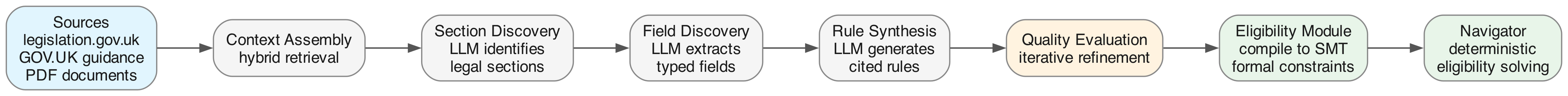}}

Every generated rule carries provenance linking it to specific source
material. Source items are tagged with structured references (e.g.,
\texttt{{[}BNA1981\#Schedule1/P1.1{]}}) in the LLM's context, and the
LLM is instructed to cite these tags on each generated criterion and
field. A verification step then resolves these citations against the
source material, creating anchors with document IDs, section paths, and
direct quotes:

\textbf{Table 3: Source Provenance Chain}

{\def\LTcaptype{none} % do not increment counter
\begin{longtable}[]{@{}
  >{\raggedright\arraybackslash}p{(\linewidth - 6\tabcolsep) * \real{0.2234}}
  >{\raggedright\arraybackslash}p{(\linewidth - 6\tabcolsep) * \real{0.2660}}
  >{\raggedright\arraybackslash}p{(\linewidth - 6\tabcolsep) * \real{0.2021}}
  >{\raggedright\arraybackslash}p{(\linewidth - 6\tabcolsep) * \real{0.3085}}@{}}
\toprule\noalign{}
\begin{minipage}[b]{\linewidth}\raggedright
Source Type
\end{minipage} & \begin{minipage}[b]{\linewidth}\raggedright
Endpoint
\end{minipage} & \begin{minipage}[b]{\linewidth}\raggedright
Format
\end{minipage} & \begin{minipage}[b]{\linewidth}\raggedright
Citation Example
\end{minipage} \\
\midrule\noalign{}
\endhead
\bottomrule\noalign{}
\endlastfoot
Primary legislation & legislation.gov.uk API & Structured markup &
\texttt{BNA1981\#Schedule1/P1.1} \\
Policy guidance & GOV.UK Content API & Structured API &
\texttt{GOVUK\#english-lang/part-2} \\
Form guidance & PDF parser & Structured text &
\texttt{form-an\#section-4/para-3} \\
\end{longtable}
}

Citations in all three source types are verified by direct resolution
from the LLM-cited references.

When the LLM fails to cite a source (or cites an invalid reference), the
system falls back to embedding-based semantic similarity matching as a
secondary mechanism. This multi-stage citation verification with
semantic fallback provides higher fidelity than post-hoc similarity
matching alone, because the LLM that generated the rule knows which
source material it used.

This provenance chain is the foundation of auditability: any
determination can be traced through the rule that produced it to the
specific source clause that authorises it.

\subsection{5.3 Phase 2: The Eligibility
Module}\label{phase-2-the-eligibility-module}

The authoring model emits rules in a constrained DSL that is parsed and
compiled to formal constraints using an SMT-based constraint evaluation
engine. The DSL is never executed directly.

The critical property of the compilation is how alternative legal routes
are represented. Each eligibility route becomes a formally independent
branch:

\textbf{Figure 2: Route Independence in the Eligibility Module}

\pandocbounded{\includegraphics[keepaspectratio,alt={Route Independence in the Eligibility Module}]{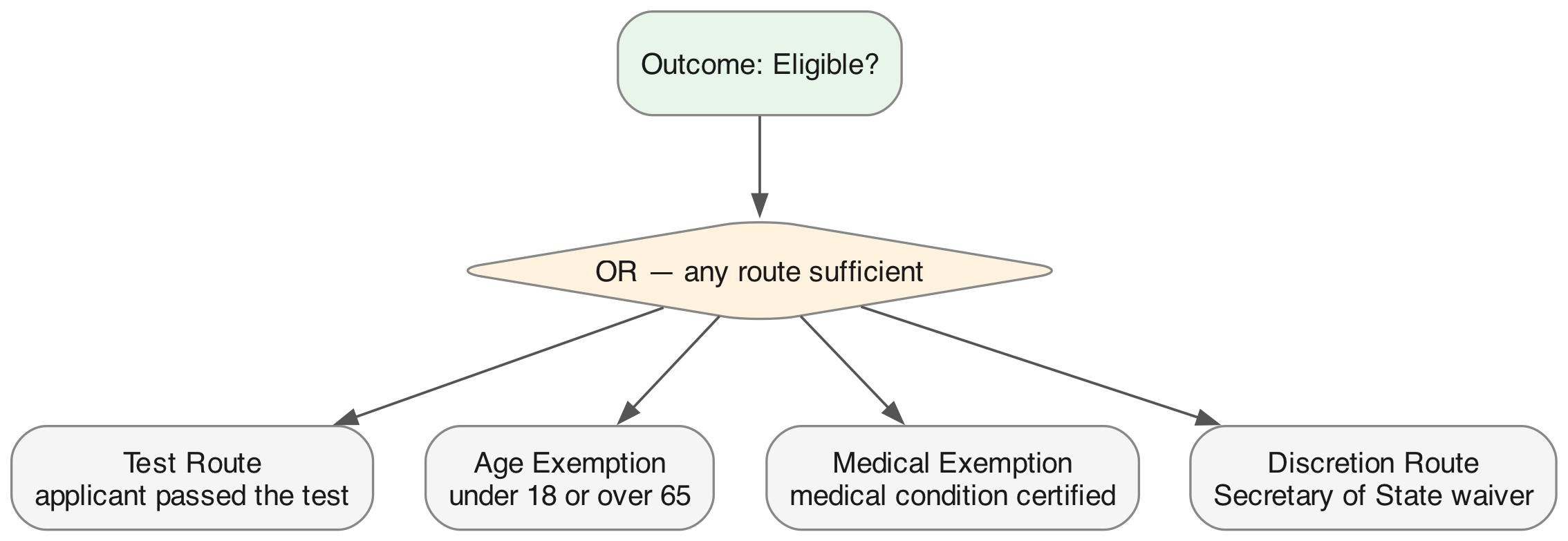}}

Each branch is a boolean expression evaluated independently. If
\texttt{medical\_exemption} evaluates to \texttt{True}, the OR is
satisfied regardless of \texttt{test\_route}. This is guaranteed by the
execution semantics: it is not a statistical property and not dependent
on model version or sampling parameters. The guarantee is over execution
given the compiled formula; whether the authored formula faithfully
represents the source legislation is the Level 2 problem (Section 5.5).

\subsection{5.4 Why the Eligibility Module Excludes These Errors by
Construction}\label{why-the-eligibility-module-excludes-these-errors-by-construction}

The LLM failures in our benchmark fall into two categories, both of
which are excluded by the execution semantics of the Module. To make
this precise, we state the compilation formally.

Let \(\mathcal{F} = \{f_1, \ldots, f_m\}\) be the set of typed applicant
fields, and let \(\sigma : \mathcal{F} \to \mathcal{V}\) be an applicant
assignment (each field mapped to a typed value). A \textbf{rule} is a
quantifier-free formula \(\phi\) in the theory of bitvectors, linear
integer/real arithmetic, and uninterpreted functions over
\(\mathcal{F}\). A \textbf{rule bundle} is a tuple
\(\mathcal{R} = \langle \mathcal{G}, \mathcal{O} \rangle\) where
\(\mathcal{G} = \{G_1, \ldots, G_k\}\) is a set of rule groups, each
\(G_i = \{\phi_{i,1}, \ldots, \phi_{i,n_i}\}\) representing independent
legal routes within a requirement, and \(\mathcal{O}\) is an outcome
combinator.

\textbf{Category 1 --- Exemption anchoring (formally excluded).} Within
a rule group \(G_i\), routes are combined disjunctively:

\[
[\![ G_i ]\!]_\sigma \;=\; \bigvee_{j=1}^{n_i} [\![ \phi_{i,j} ]\!]_\sigma
\]

The semantics are: \([\![ G_i ]\!]_\sigma = \top\) iff \emph{any} route
satisfies the applicant assignment. There is no weighting, no ordering,
and no context-dependent attenuation of later disjuncts by earlier ones.
An LLM's learned approximation of this disjunction is not guaranteed to
exhibit this property; the SMT-evaluated formula is, by the soundness of
the decision procedure, evaluated over the authored formula. Whether
that authored formula faithfully captures the legislator's intent is a
separate question, addressed at Level 2 (Section 5.5).

\textbf{Category 2 --- Exception chain collapse (formally excluded).} A
multi-level exception chain of the form

\begin{quote}
\emph{A is required, UNLESS B applies, UNLESS C revokes B, UNLESS D
overrides everything}
\end{quote}

compiles to the material implication

\[
\underbrace{
  \neg\Big[\,
    \underbrace{(B \land \neg C)}_{\text{exemption, not revoked}}
    \;\lor\;
    \underbrace{D}_{\text{override}}
  \,\Big]
}_{\text{no exemption or override active}}
\;\Rightarrow\;
A
\]

or equivalently \((B \land \neg C) \lor D \lor A\). This is a
closed-form Boolean expression over the applicant fields. Evaluating it
against a fixed applicant assignment and a compiled rule bundle is a
deterministic computation; the result is a Boolean truth value with no
variance over re-evaluation, temperature, or model version.

\textbf{Contrast with LLM inference.} An LLM computes

\[
p_\theta(\text{eligible} \mid \text{legislation},\, \sigma) \;\approx\; \mathbb{1}[\phi_{\text{gt}}(\sigma) = \top]
\]

where \(\phi_{\text{gt}}\) is the ground-truth formula. The left-hand
side is a learned distribution over tokens; the approximation is
empirically good on shallow compositional tasks and empirically poor on
three-level exception chains (Section 6). The right-hand side --- what
the engine computes --- is the indicator function itself. The failure
patterns in Section 4 are failures of the approximation, and they are
excluded by construction whenever the evaluator operates directly on
\(\phi_{\text{gt}}\). The guarantee covers Level 3 (execution) only;
whether the authored \(\phi\) faithfully represents \(\phi_{\text{gt}}\)
is the Level 2 problem (Section 5.5, Section 7).

\subsection{5.5 What the Eligibility Module Guarantees - and What It
Does
Not}\label{what-the-eligibility-module-guarantees---and-what-it-does-not}

The system guarantees correctness of execution given a correct
specification; it does not guarantee correctness of the specification
itself.

Stated more fully: the system provides deterministic guarantees at one
level only --- \textbf{correct execution of formalised rules}. It makes
no claim about two upstream steps:

\textbf{Level 1 - Source text retrieval.} The system retrieves
authoritative source material using source connectors and retrieval
pipelines. Errors here - missing sections, retrieval gaps - will produce
incorrectly grounded rules. The system does not guarantee that all
relevant source text was found and correctly parsed.

\textbf{Level 2 - Rule formalisation.} LLMs generate rules from source
material. An incorrectly formalised rule produces incorrect
determinations, even if the execution is formally correct. This is the
``garbage in, garbage out'' property stated precisely: deterministic
execution of incorrectly formalised rules produces incorrect but
deterministic results. The system addresses this through test-driven
validation: subject matter experts define test cases covering golden
paths, edge cases, and corner cases, and authored rules must pass these
test suites before deployment (Section 7.3). Rules can also be reviewed
by human experts, but the primary quality gate is automated test-case
validation, not manual review.

\textbf{Level 3 - Rule execution.} Once rules are correctly formalised,
the engine guarantees correct execution with mathematically defined
semantics. The failure patterns documented in Section 4 - exemption
anchoring and exception chain collapse - are excluded by the execution
semantics at this level.

This distinction matters for regulatory defensibility. The provenance
chain (Section 9.1) supports audit of Levels 1 and 2; Level 3 is
guaranteed by the execution semantics. Execution correctness is a
necessary precondition for trust in the overall system: if the execution
layer itself could introduce errors, no amount of rule quality
engineering would produce reliable determinations. By solving Level 3
first, the system reduces the trust problem to a single surface --- rule
formalisation quality --- which is addressed through iterative synthesis
refinement (Section 7.1) and test-driven validation against SME-defined
test suites (Section 7.3).

\begin{center}\rule{0.5\linewidth}{0.5pt}\end{center}

\section{6. Accuracy Benchmark: Eligibility Module vs Eight
LLMs}\label{accuracy-benchmark-eligibility-module-vs-eight-llms}

\subsection{6.1 Methodology}\label{methodology}

Both systems receive identical inputs: the same legislation text
(markdown-formatted) and the same structured applicant data (field
key-value pairs). The comparison isolates the reasoning engine.
Everything else is held constant.

\begin{itemize}
\tightlist
\item
  \textbf{System A (Eligibility Module):} Gold-standard rule fixtures
  compiled to formal SMT constraints. The navigator evaluates
  constraints against structured applicant data and returns a
  deterministic eligible/ineligible determination.
\item
  \textbf{System B (LLM baseline):} The same source legislation is
  provided to each LLM along with the same structured applicant data.
  The LLM determines eligibility and responds with
  \texttt{\{"eligible":\ true/false,\ "reasoning":\ "..."\}}.
\end{itemize}

\textbf{Models tested:}

{\def\LTcaptype{none} % do not increment counter
\begin{longtable}[]{@{}llll@{}}
\toprule\noalign{}
Model & Provider & Class & Cost Tier \\
\midrule\noalign{}
\endhead
\bottomrule\noalign{}
\endlastfoot
Claude Opus 4.6 & Anthropic & Frontier & Premium \\
Claude Sonnet 4.6 & Anthropic & Frontier & Mid \\
GPT-5.4 & OpenAI & Frontier & Premium \\
GPT-5.3 & OpenAI & Production & Mid \\
GPT-5-mini & OpenAI & Production & Low \\
GPT-5-nano & OpenAI & Production & Lowest \\
Claude Opus 4.7 & Anthropic & Frontier & Premium \\
GPT-4.1-mini & OpenAI & Production & Low \\
\end{longtable}
}

Claude Opus 4.7 and GPT-4.1-mini were added in the April 2026 v3.8
replication (Opus 4.7 in Tables 6, 8a, 8c and throughout §6.10;
GPT-4.1-mini in Tables 7, 8a, 8b); they are not part of the original
March 2026 six-model selection. GPT-4.1-mini serves as a cost-tier
baseline for the construction domain.

\textbf{Evaluation coverage.} Not all models were evaluated on all
domains. The following matrix shows the scope of each model's
evaluation:

{\def\LTcaptype{none} % do not increment counter
\begin{longtable}[]{@{}
  >{\raggedright\arraybackslash}p{(\linewidth - 10\tabcolsep) * \real{0.2059}}
  >{\centering\arraybackslash}p{(\linewidth - 10\tabcolsep) * \real{0.1471}}
  >{\centering\arraybackslash}p{(\linewidth - 10\tabcolsep) * \real{0.1471}}
  >{\centering\arraybackslash}p{(\linewidth - 10\tabcolsep) * \real{0.1471}}
  >{\centering\arraybackslash}p{(\linewidth - 10\tabcolsep) * \real{0.1471}}
  >{\raggedright\arraybackslash}p{(\linewidth - 10\tabcolsep) * \real{0.2059}}@{}}
\toprule\noalign{}
\begin{minipage}[b]{\linewidth}\raggedright
Model
\end{minipage} & \begin{minipage}[b]{\linewidth}\centering
life\_uk (56)
\end{minipage} & \begin{minipage}[b]{\linewidth}\centering
english\_language (43)
\end{minipage} & \begin{minipage}[b]{\linewidth}\centering
spacecraft (68)
\end{minipage} & \begin{minipage}[b]{\linewidth}\centering
construction\_car (58)
\end{minipage} & \begin{minipage}[b]{\linewidth}\raggedright
Notes
\end{minipage} \\
\midrule\noalign{}
\endhead
\bottomrule\noalign{}
\endlastfoot
Claude Opus 4.6 & ✓ & ✓ & ✓ & ✓ (v3.8 replication) & + robustness
(N=10) \\
Claude Sonnet 4.6 & ✓ & ✓ & ✓ & ✓ (v3.8 replication) & + §6.4.1
adversarial \\
GPT-5.4 & ✓ & ✓ & ✓ & ✓ & + enhanced prompt; + §6.4.1 adversarial
(default and low reasoning); + n=11 instrumented replication \\
GPT-5.3 & --- & --- & --- & 11-scenario subset & production reference \\
GPT-5-mini & ✓ & ✓ & ✓ & --- & \\
GPT-5-nano & --- & ✓ & 48-scenario baseline & --- & cost reference \\
Claude Opus 4.7 & --- & --- & ✓ & ✓ & + §6.4.1 adversarial; v3.8
replication only \\
GPT-4.1-mini & --- & --- & --- & ✓ & construction cost-tier reference \\
\end{longtable}
}

Results reported below should be read against this matrix. Aggregate
claims refer only to evaluations actually conducted.

\textbf{Model selection.} The original six-model selection (March 2026)
spans three attributes: (i) frontier versus production tier, (ii) the
two providers our organisation had API access to during the evaluation
window (Anthropic and OpenAI), and (iii) a range of cost tiers
representing realistic deployment choices. Models from other providers
(Google Gemini, DeepSeek, Meta Llama, Mistral) were not included in this
pass due to access and budget constraints; broader provider coverage is
pre-registered for replication at N=66 (Section 6.9). The selection was
fixed before any results were computed and was not adjusted based on
intermediate findings. Claude Opus 4.7 and GPT-4.1-mini were added
later, at v3.8 replication time --- Opus 4.7 as the then-strongest
Anthropic model and GPT-4.1-mini as a cost-tier reference for the
construction domain --- which post-dates this original fixed selection.

\textbf{LLM configuration:} Temperature 0 where supported (Claude
models; GPT-5 family does not expose this parameter --- sampling is
controlled internally by the model's reasoning pipeline). The §6.7
robustness baseline was additionally run at T=0.3 as a
temperature-sensitivity control; Table 9 shows the failure set is
identical at T=0.3 and T=0. Runs per scenario: 3 (majority vote;
robustness analysis tests N=10). Prompt: generic legal assessment, no
special instructions about exemptions or OR-branching.

\textbf{Prompt variants explored.} Two to three prompt variants were
explored informally during initial development to confirm that the
observed failure pattern was not an artefact of wording --- for example,
whether instructing the LLM to ``think step by step'' or to ``consider
all exemption routes before deciding'' closed the gap on exception-chain
scenarios. These variants did not materially change the pattern and were
not logged systematically. Systematic prompt comparison across three
variants per scenario is pre-registered for the N=66 replication
(Section 6.9). The enhanced prompt analysed in Section 6.7 is a fourth
variant specifically targeting the failure pattern.

\textbf{Benchmark philosophy.} The goal is not to find the best possible
LLM score through bespoke prompt engineering. It is to test whether a
general-purpose model reliably preserves exception structure when
reading authoritative rules in the way such systems are actually
deployed - with a reasonable, generic instruction. This is a different
question from ``what is the ceiling of LLM performance on this task?''
The robustness analysis (Section 6.7) directly addresses whether the gap
is a prompt engineering problem or a deeper reliability issue.

\textbf{Domain disclosure.} The four benchmark domains span different
levels of source authenticity:

{\def\LTcaptype{none} % do not increment counter
\begin{longtable}[]{@{}
  >{\raggedright\arraybackslash}p{(\linewidth - 4\tabcolsep) * \real{0.2162}}
  >{\raggedright\arraybackslash}p{(\linewidth - 4\tabcolsep) * \real{0.4324}}
  >{\raggedright\arraybackslash}p{(\linewidth - 4\tabcolsep) * \real{0.3514}}@{}}
\toprule\noalign{}
\begin{minipage}[b]{\linewidth}\raggedright
Domain
\end{minipage} & \begin{minipage}[b]{\linewidth}\raggedright
Source material
\end{minipage} & \begin{minipage}[b]{\linewidth}\raggedright
Authenticity
\end{minipage} \\
\midrule\noalign{}
\endhead
\bottomrule\noalign{}
\endlastfoot
\texttt{life\_uk} & British Nationality Act 1981, Schedule 1 + Home
Office guidance & Real UK legislation \\
\texttt{english\_language} & Form AN guidance + English language policy
& Real UK legislation \\
\texttt{spacecraft} & Spacecraft Crew Certification Act (synthetic
statute) & Synthetic, modelled on UK legislative structure \\
\texttt{construction\_car} & CAR policy defect exclusion endorsement
(synthetic wording) & Synthetic, modelled on London market DE3/DE5
clauses \\
\end{longtable}
}

All models evaluated on the \texttt{life\_uk} section (see the coverage
matrix) achieve 100\% (depth-1 combinatorial boolean logic); this
section is reported here for completeness but not charted separately.
All frontier models evaluated on \texttt{english\_language} also achieve
100\% (depth-2 multi-route logic); production-tier GPT-5-nano scores
48.8\% on the same section (Table 5). These two sections establish a
baseline: the failure pattern is specific to exception chain depth, not
general to legal reasoning.

\textbf{Note on immigration sections.} The life\_uk and
english\_language sections cover two specific isolated requirements
within UK naturalisation eligibility. The full naturalisation
determination involves additional sections with exception chain
structures of equivalent or greater complexity to the spacecraft
section; those are not included here.

\textbf{Fixture independence.} The rule fixtures used by the Eligibility
Module were authored independently of the LLM baseline evaluations and
fixed prior to benchmarking. LLMs were not used to generate the
ground-truth rules against which they are evaluated.

\textbf{Scope note on construction insurance materials.} The
\texttt{construction\_car} benchmark scenarios and their expected
outcomes were authored by the primary author (a non-lawyer) based on
published DE3/DE5 clause wordings and industry commentary. They have not
been independently validated by a qualified construction-insurance
professional and should therefore be read as author-constructed
illustrative examples of the exception-chain structure in this domain
rather than as professionally validated case studies. The immigration
benchmarks (\texttt{life\_uk} and \texttt{english\_language}) were
authored by L.D., an Immigration Solicitor and accredited Senior
Caseworker (see Author Contributions).

\textbf{Note on repository scope.} The public benchmark repository (see
Data and Code Availability) additionally contains a
\texttt{benefits\_entitlement} domain that is not analysed in this
paper. It is included in the repository for completeness and future work
but is outside the scope of this paper; all results, tables and findings
reported here cover only the four domains listed above.

\textbf{Benchmark intent.} The benchmark is designed to exercise
nested-exception-chain structures that occur in real legislation and
commercial contracts, not synthetic structures contrived solely to
confuse LLMs. The British Nationality Act 1981 Schedule 1 and JCT 2016
DE3/DE5 insurance wordings both contain multi-level UNLESS chains in
practice; we deliberately selected and constructed scenarios that
exercise this structure rather than average-case queries against these
sources. The finding is therefore about how LLMs handle a structure that
recurs in high-stakes legal and insurance texts, not about whether we
can synthesise inputs that confuse them. Typical-case performance on the
same sources is outside the scope of this paper.

\subsection{6.2 Scenario Design}\label{scenario-design}

\textbf{Table 4: Benchmark Scenario Coverage}

{\def\LTcaptype{none} % do not increment counter
\begin{longtable}[]{@{}lccc@{}}
\toprule\noalign{}
Section & Scenarios & Fields & Difficulty \\
\midrule\noalign{}
\endhead
\bottomrule\noalign{}
\endlastfoot
life\_uk & 56 & 4 & Low \\
english\_language & 43 & 12 & Medium \\
spacecraft & 68 & 11 & High \\
construction\_car & 58 & 14 & High \\
\textbf{Total} & \textbf{225} & & \\
\end{longtable}
}

\textbf{Key patterns by section.} - \emph{life\_uk:} combinatorial --- 3
booleans × 7 ages. - \emph{english\_language:} Canada trap, SELT edge
cases, near-misses, multi-field interactions. - \emph{spacecraft:}
three-level exception chain, nested IMPLIES, UNSAT early termination,
adversarial. - \emph{construction\_car:} DE3/DE5 defect exclusion,
access damage carve-back, enhanced cover chain, pioneer override, known
defect limitation.

\subsection{6.3 Results}\label{results}

In all tables below, accuracy figures for the Eligibility Module refer
to agreement with the benchmark's formal rule fixtures --- the authored
rule specifications used as the benchmark's ground truth. LLM accuracy
is measured against the same fixtures.

\textbf{Table 5: English Language: 43 Scenarios, 12 Fields (Medium
Difficulty)}

All accuracy figures report Wilson 95\% confidence intervals.

{\def\LTcaptype{none} % do not increment counter
\begin{longtable}[]{@{}
  >{\raggedright\arraybackslash}p{(\linewidth - 10\tabcolsep) * \real{0.1186}}
  >{\centering\arraybackslash}p{(\linewidth - 10\tabcolsep) * \real{0.1695}}
  >{\centering\arraybackslash}p{(\linewidth - 10\tabcolsep) * \real{0.2542}}
  >{\centering\arraybackslash}p{(\linewidth - 10\tabcolsep) * \real{0.1695}}
  >{\centering\arraybackslash}p{(\linewidth - 10\tabcolsep) * \real{0.2203}}
  >{\centering\arraybackslash}p{(\linewidth - 10\tabcolsep) * \real{0.0678}}@{}}
\toprule\noalign{}
\begin{minipage}[b]{\linewidth}\raggedright
Model
\end{minipage} & \begin{minipage}[b]{\linewidth}\centering
Accuracy
\end{minipage} & \begin{minipage}[b]{\linewidth}\centering
95\% Wilson CI
\end{minipage} & \begin{minipage}[b]{\linewidth}\centering
Boundary
\end{minipage} & \begin{minipage}[b]{\linewidth}\centering
Consistency
\end{minipage} & \begin{minipage}[b]{\linewidth}\centering
FN
\end{minipage} \\
\midrule\noalign{}
\endhead
\bottomrule\noalign{}
\endlastfoot
\textbf{Eligibility Module} & \textbf{43/43 (100\%)} &
\textbf{{[}91.8\%, 100\%{]}} & \textbf{7/7 (100\%)} & \textbf{43/43
(100\%)} & \textbf{0} \\
Claude Opus 4.6 & 43/43 (100\%) & {[}91.8\%, 100\%{]} & 7/7 (100\%) &
43/43 (100\%) & 0 \\
Claude Sonnet 4.6 & 43/43 (100\%) & {[}91.8\%, 100\%{]} & 7/7 (100\%) &
43/43 (100\%) & 0 \\
GPT-5.4 & 43/43 (100\%) & {[}91.8\%, 100\%{]} & 7/7 (100\%) & 43/43
(100\%) & 0 \\
GPT-5-mini & 42/43 (97.7\%) & {[}87.9\%, 99.6\%{]} & 7/7 (100\%) & 39/43
(91\%) & 1 \\
GPT-5-nano & 21/43 (48.8\%) & {[}34.6\%, 63.2\%{]} & 6/7 (86\%) & 40/43
(93\%) & 22 \\
\end{longtable}
}

\textbf{Figure 3: Spacecraft accuracy with 95\% Wilson confidence
intervals.}

\pandocbounded{\includegraphics[keepaspectratio,alt={Spacecraft section forest plot: Wilson 95\% CIs across models. The Eligibility Module and GPT-5.4 sit at the ceiling (both 68/68, CI {[}94.7\%, 100\%{]}); Claude Opus and Sonnet overlap in the {[}80\%, 96\%{]} band and are not statistically distinguishable from each other; GPT-5-mini and nano trail below. Generated from Table 6 data by figures/scripts/forest\_plot\_spacecraft.py.}]{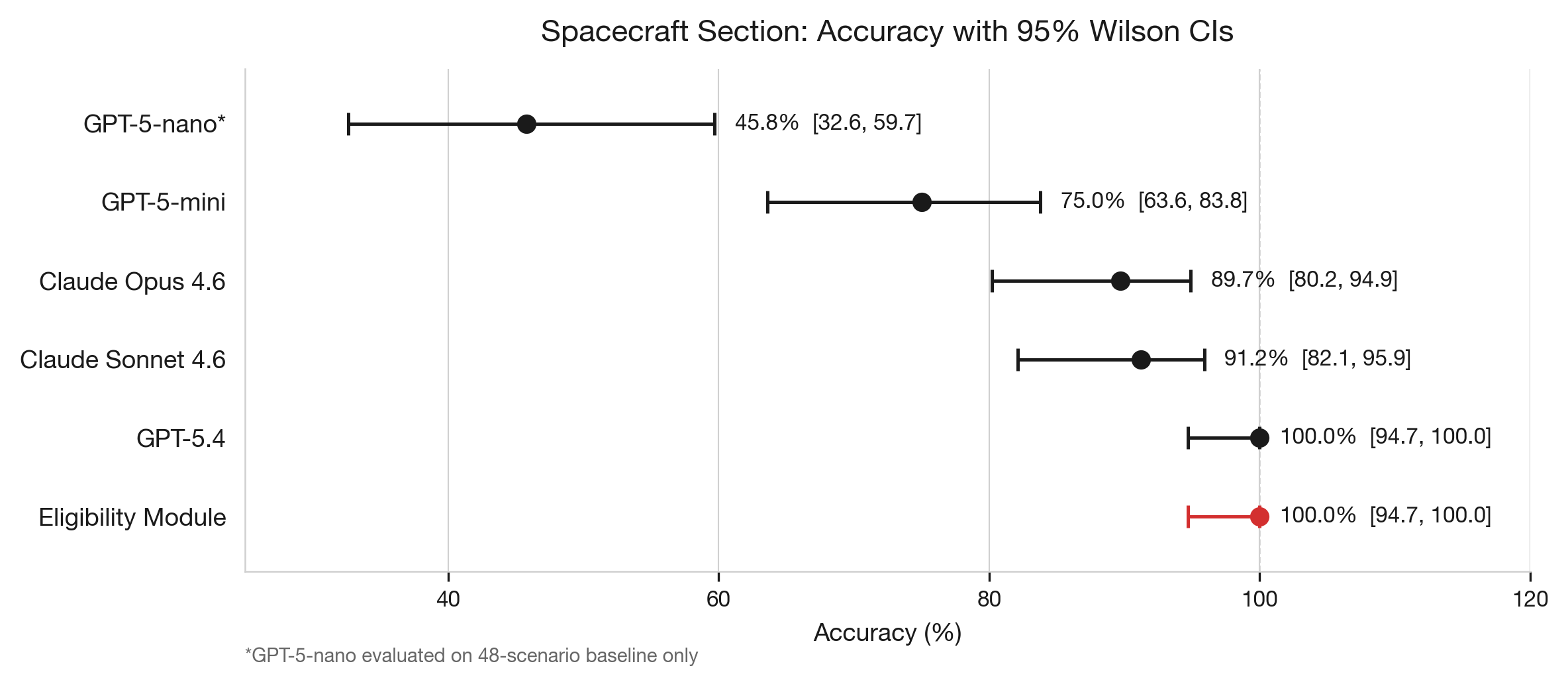}}

\textbf{Table 6: Spacecraft: 68 Scenarios, 11 Fields, Adversarial
Exception Chains (High Difficulty)}

Numbers are the v3.6 / v3.7 snapshot (March 2026); v3.8 replication
numbers are listed alongside where they differ. The shifting columns are
the central evidence for the model-drift point made in §6.5 Finding 4
and the abstract.

{\def\LTcaptype{none} % do not increment counter
\begin{longtable}[]{@{}
  >{\raggedright\arraybackslash}p{(\linewidth - 4\tabcolsep) * \real{0.4118}}
  >{\centering\arraybackslash}p{(\linewidth - 4\tabcolsep) * \real{0.1569}}
  >{\centering\arraybackslash}p{(\linewidth - 4\tabcolsep) * \real{0.4314}}@{}}
\toprule\noalign{}
\begin{minipage}[b]{\linewidth}\raggedright
Model
\end{minipage} & \begin{minipage}[b]{\linewidth}\centering
March 2026 (v3.7)
\end{minipage} & \begin{minipage}[b]{\linewidth}\centering
April 2026 (v3.8 replication, same prompt + harness)
\end{minipage} \\
\midrule\noalign{}
\endhead
\bottomrule\noalign{}
\endlastfoot
\textbf{Eligibility Module} & \textbf{68/68 (100\%) {[}94.7--100{]}} &
\textbf{68/68 (100\%)} (deterministic) \\
GPT-5.4 & 68/68 (100\%) & not re-tested (was already at 100\%) \\
Claude Opus 4.7 (Anthropic's strongest at evaluation, Apr 2026) & not in
v3.7 & \textbf{68/68 (100\%)} {[}94.7--100{]} \\
Claude Opus 4.6 & 61/68 (89.7\%) {[}80.2--94.9{]} & \textbf{67/68
(98.5\%)} {[}92.1--99.7{]} --- 6 fewer failures \\
Claude Sonnet 4.6 & 62/68 (91.2\%) {[}82.1--95.9{]} & 60/68 (88.2\%)
{[}78.5--93.9{]} --- 2 more failures \\
GPT-5-mini & 51/68 (75.0\%) {[}63.6--83.8{]} & not re-tested \\
GPT-5-nano* & 22/48 (45.8\%) {[}32.6--59.7{]} & not re-tested \\
\end{longtable}
}

\emph{*GPT-5-nano was evaluated on the 48-scenario baseline suite only
(pre-adversarial expansion) and is included as a cost-scaled reference,
not as a realistic deployment choice.}

Replication harness: \texttt{tools/replication\_run.py} in the public
repo, run via the paper's exact \texttt{\_build\_prompt} format.
Per-scenario JSONs at \texttt{docs/replication/A*.json}. The Opus 4.6
89.7\% → 98.5\% drift between March and April invalidates the §6.7 R1
70-trial-on-7-failing-scenarios result as currently written; see §6.7 R1
caveat.

\subsection{6.4 Construction Insurance: DE3/DE5 Defect Exclusion (58
Scenarios)}\label{construction-insurance-de3de5-defect-exclusion-58-scenarios}

The construction insurance benchmark extends the evaluation into a
domain with commercial relevance: London market Construction All Risks
(CAR) policy wording. The test fixture is modelled on DE3/DE5 defect
exclusion clauses used for major infrastructure projects.

\textbf{The policy logic.} A CAR policy covers physical loss or damage
to the Works. The defect exclusion creates a five-level exception chain
(a base exclusion modified by five successive levels of exception;
scenario labels such as ``depth-6'' count the base as level 1):

{\def\LTcaptype{none} % do not increment counter
\begin{longtable}[]{@{}
  >{\raggedright\arraybackslash}p{(\linewidth - 4\tabcolsep) * \real{0.2174}}
  >{\raggedright\arraybackslash}p{(\linewidth - 4\tabcolsep) * \real{0.3478}}
  >{\raggedright\arraybackslash}p{(\linewidth - 4\tabcolsep) * \real{0.4348}}@{}}
\toprule\noalign{}
\begin{minipage}[b]{\linewidth}\raggedright
Level
\end{minipage} & \begin{minipage}[b]{\linewidth}\raggedright
Rule
\end{minipage} & \begin{minipage}[b]{\linewidth}\raggedright
Plain English
\end{minipage} \\
\midrule\noalign{}
\endhead
\bottomrule\noalign{}
\endlastfoot
\textbf{Exclusion} & Rectification cost absolutely excluded & Cost of
fixing the defective part: never covered \\
\textbf{Carve-back} & Resultant damage to other property covered &
Non-defective property damaged by the failure: covered \\
\textbf{Re-exclusion} & Access damage excluded & Property damaged just
to reach the defect: not covered \\
\textbf{Enhanced cover} & Projects \textgreater= £100M: access damage
reinstated & Large projects get access damage back\ldots{} \\
\textbf{Design limit} & \ldots except for design defects & \ldots but
not if the defect was in the design \\
\textbf{Pioneer override} & Projects \textgreater= £500M: design limit
removed & Mega-projects override the design limitation \\
\end{longtable}
}

\textbf{Table 7: Construction Insurance Exception Chain (access damage
scenarios)}

Cells reflect the v3.7 snapshot (March 2026). The ``Access, £500 M,
design'' cell where GPT-5.4 was reported as ✗ does not replicate in
April 2026 --- current GPT-5.4 returns the correct answer (covered) on
this scenario; see Table 8a v3.8 column. The cell is preserved here as
the v3.7 historical record because the model-drift evidence is itself
part of the paper's argument.

{\def\LTcaptype{none} % do not increment counter
\begin{longtable}[]{@{}
  >{\raggedright\arraybackslash}p{(\linewidth - 10\tabcolsep) * \real{0.3529}}
  >{\centering\arraybackslash}p{(\linewidth - 10\tabcolsep) * \real{0.0706}}
  >{\centering\arraybackslash}p{(\linewidth - 10\tabcolsep) * \real{0.1176}}
  >{\centering\arraybackslash}p{(\linewidth - 10\tabcolsep) * \real{0.2353}}
  >{\centering\arraybackslash}p{(\linewidth - 10\tabcolsep) * \real{0.0824}}
  >{\centering\arraybackslash}p{(\linewidth - 10\tabcolsep) * \real{0.1412}}@{}}
\toprule\noalign{}
\begin{minipage}[b]{\linewidth}\raggedright
Scenario
\end{minipage} & \begin{minipage}[b]{\linewidth}\centering
Project
\end{minipage} & \begin{minipage}[b]{\linewidth}\centering
Defect
\end{minipage} & \begin{minipage}[b]{\linewidth}\centering
Eligibility Module
\end{minipage} & \begin{minipage}[b]{\linewidth}\centering
GPT-5.4 (v3.7)
\end{minipage} & \begin{minipage}[b]{\linewidth}\centering
GPT-4.1-mini
\end{minipage} \\
\midrule\noalign{}
\endhead
\bottomrule\noalign{}
\endlastfoot
Access, £100M, workmanship & £100M & workmanship & COVERED & ✓ & ✗ \\
Access, £200M, materials & £200M & materials & COVERED & ✓ & ✗ \\
Access, £200M, design & £200M & design & NOT COVERED & ✓ & ✓ \\
Access, £500M, design & £500M & design & COVERED & ✗ (v3.7) → ✓ (v3.8) &
✗ \\
Access, £800M, design & £800M & design & COVERED & ✓ & ✗ \\
\end{longtable}
}

\textbf{Figure 4: Construction insurance accuracy with 95\% Wilson
confidence intervals.}

\pandocbounded{\includegraphics[keepaspectratio,alt={Construction insurance forest plot: Wilson 95\% CIs across the three models evaluated on the full 58-scenario suite. The Eligibility Module sits at the 100\% ceiling (CI {[}93.8\%, 100\%{]}); GPT-5.4 at 96.6\% (CI {[}88.3\%, 99.0\%{]}) is distinguishable from GPT-4.1-mini at 79.3\% (CI {[}67.2\%, 87.7\%{]}). Generated from Table 8a data by figures/scripts/forest\_plot\_construction.py.}]{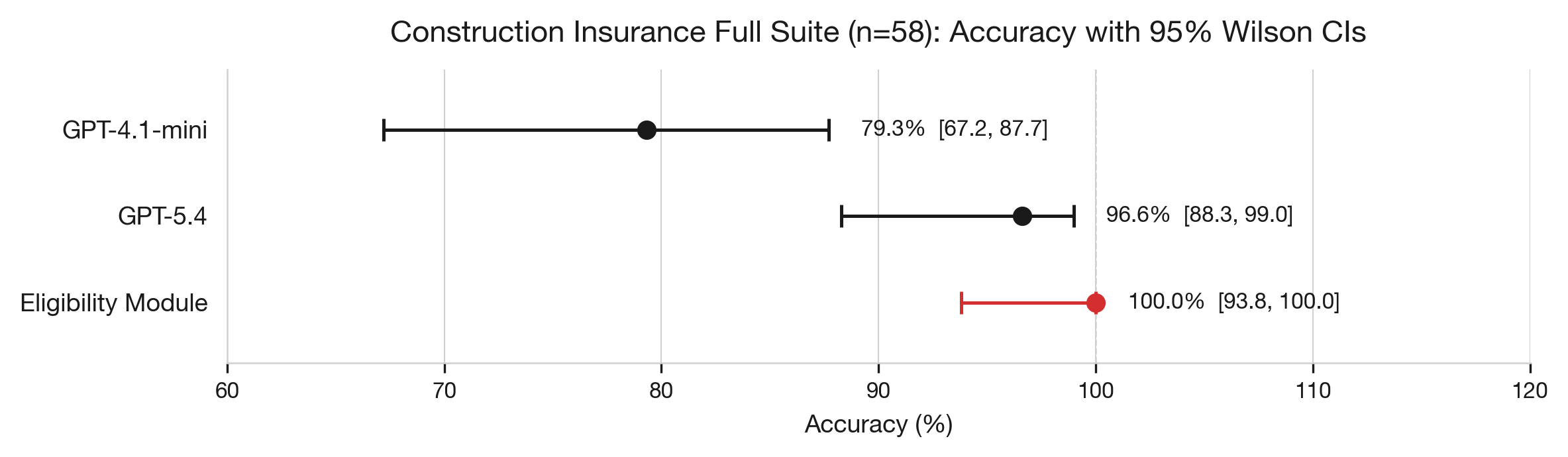}}

\textbf{Table 8a: Construction Insurance --- Full Suite (March 2026 vs
April 2026 replication)}

Primary results on the 58-scenario construction insurance benchmark.
v3.7 numbers (March 2026 snapshot) shown alongside v3.8 replication
(April 2026, same prompt and harness, current model alias). Wilson 95\%
CIs in brackets.

{\def\LTcaptype{none} % do not increment counter
\begin{longtable}[]{@{}
  >{\raggedright\arraybackslash}p{(\linewidth - 4\tabcolsep) * \real{0.2985}}
  >{\raggedright\arraybackslash}p{(\linewidth - 4\tabcolsep) * \real{0.4030}}
  >{\raggedright\arraybackslash}p{(\linewidth - 4\tabcolsep) * \real{0.2985}}@{}}
\toprule\noalign{}
\begin{minipage}[b]{\linewidth}\raggedright
Model
\end{minipage} & \begin{minipage}[b]{\linewidth}\raggedright
March 2026 (v3.7)
\end{minipage} & \begin{minipage}[b]{\linewidth}\raggedright
April 2026 (v3.8 replication, full 58-suite + 16 early scenarios from
the §6.9 exception-chain expansion set)
\end{minipage} \\
\midrule\noalign{}
\endhead
\bottomrule\noalign{}
\endlastfoot
\textbf{Eligibility Module} & \textbf{58/58 (100\%) {[}93.8--100.0{]}} &
\textbf{74/74 (100\%)} {[}95.1--100{]} (deterministic) \\
GPT-5.4 & 56/58 (96.6\%) {[}88.3--99.0{]} & \textbf{74/74 (100\%)}
{[}95.1--100{]} --- closed silently between March and April \\
Claude Opus 4.7 (Anthropic's strongest at evaluation, Apr 2026) & not
tested in v3.7 & 74/74 (100\%) {[}95.1--100{]} \\
Claude Opus 4.6 & not tested in v3.7 & 74/74 (100\%) {[}95.1--100{]} \\
Claude Sonnet 4.6 & not tested in v3.7 & 72/74 (97.3\%) {[}90.7--99.3{]}
--- fails on \texttt{access\_100m\_design\_not\_covered},
\texttt{surface\_eligible\_design\_access\_blocks} \\
GPT-4.1-mini & 46/58 (79.3\%) {[}67.2--87.7{]} & 65/74 (87.8\%)
{[}78.5--93.5{]} --- direction holds; magnitude shifted; still fails on
\texttt{neg\_*} (negation\_stacking), \texttt{surface\_*}
(contradictory\_cue), and \texttt{ultimate\_trap\_1} \\
\end{longtable}
}

\emph{GPT-5.3 was the cost-tier production reference in v3.7 (claimed
7/11 on the n=11 exception-chain subset). The model alias was deprecated
by OpenAI between March and April 2026
(\texttt{NotFoundError:\ model\ gpt-5.3\ does\ not\ exist}). Replication
impossible.}

The v3.7 paper's lead frontier-LLM failure case (GPT-5.4 missing 2/58
construction scenarios) does not replicate in v3.8. The v3.4 / v3.7
narrative \emph{``no frontier model achieves 100\% across all four paper
domains''} must be qualified: as of April 2026, GPT-5.4, Opus 4.7, and
Opus 4.6 all reach 100\% on the v3.7 58-scenario construction benchmark
using the same paper-prompt format. Sonnet 4.6 still fails 2/74;
GPT-4.1-mini still fails 9/74; the cost-tier and a small set of mid-tier
failure modes remain visible. \textbf{The v3.7 paper suite no longer
differentiates the engine from the strongest frontier configurations.
That is the moving-boundary problem in miniature: the benchmark stood
still and the models moved --- in this instance toward the ceiling,
silently, under unchanged aliases.} §6.4.1 below documents the v3.8
adversarial extension, where the gap re-opens at deeper composition.

Replication artefacts:
\texttt{confidently-wrong-benchmark/legalbench/docs/replication/A*.json}
and \texttt{STREAM\_A\_REPORT.md} for full per-scenario JSON, including
which specific scenarios each model failed on.

\textbf{Table 8b: Construction Insurance --- Exception-Chain Sub-Study
(N=11)}

Inter-model comparison on the 11-scenario
\texttt{exception\_chain}-tagged subset. Wilson 95\% CIs in brackets.
The original v3.6 / v3.7 row ``GPT-5.4 (low reasoning) --- 7/11
(63.6\%)'' is \textbf{withdrawn in v3.8} following an instrumented
replication that returned 11/11 (100\%) correct with the low
reasoning-effort configuration explicitly requested; see
\texttt{docs/r5-withdrawal-note.md} and Finding 5 (§6.5). The §6.9
pre-registered N=66 replication remains the venue for any larger-sample
reasoning-effort test.

{\def\LTcaptype{none} % do not increment counter
\begin{longtable}[]{@{}
  >{\raggedright\arraybackslash}p{(\linewidth - 4\tabcolsep) * \real{0.2778}}
  >{\raggedright\arraybackslash}p{(\linewidth - 4\tabcolsep) * \real{0.3000}}
  >{\raggedright\arraybackslash}p{(\linewidth - 4\tabcolsep) * \real{0.4222}}@{}}
\toprule\noalign{}
\begin{minipage}[b]{\linewidth}\raggedright
Model
\end{minipage} & \begin{minipage}[b]{\linewidth}\raggedright
Exception-chain (n=11)
\end{minipage} & \begin{minipage}[b]{\linewidth}\raggedright
Notes
\end{minipage} \\
\midrule\noalign{}
\endhead
\bottomrule\noalign{}
\endlastfoot
\textbf{Eligibility Module} & \textbf{11/11 (100\%) {[}74.1--100.0{]}} &
deterministic \\
GPT-5.4 (default) & 10/11 (90.9\%) {[}62.3--98.4{]} & original v3.6 /
v3.7 figure \\
GPT-5.4 (low reasoning effort, v3.8 replication) & 11/11 (100\%)
{[}74.1--100.0{]} & instrumented; cannot reproduce v3.6 / v3.7 7/11 \\
GPT-5.3 & 7/11 (63.6\%) {[}35.4--84.8{]} & production-tier baseline \\
GPT-4.1-mini & 5/11 (45.5\%) {[}21.3--72.0{]} & cost-tier baseline \\
\end{longtable}
}

GPT-5.3 scores 7/11 (63.6\%, 95\% CI {[}35.4\%, 84.8\%{]}) on the
11-scenario exception chain subset. It correctly identifies absolute
exclusions and simple carve-backs but fails on four scenarios requiring
multi-level exception reasoning, including enhanced cover reinstatement,
the pioneer override, and the depth-5 unblock. The v3.6 / v3.7 paper
additionally claimed GPT-5.4 at low reasoning effort matched this 7/11
with the same four scenarios failing --- that intra-model
reasoning-effort claim is \textbf{withdrawn in v3.8}: an instrumented
replication on the exact same 11 scenarios returned 11/11 correct with
the low reasoning-effort configuration explicitly requested, and no
committed script reproduces the original 7/11 figure (see Finding 5,
§6.5, and \texttt{docs/r5-withdrawal-note.md}). The inter-model gap
(GPT-5.3 64\% vs GPT-5.4 default 91\% on the same chain) is real and
narrow; §6.9 pre-registers an N=66 replication for the deeper
compute-dependence question.

Note: v3.5 of this paper reported GPT-5.3 at 27\% (3/11). That figure
was inflated by a configuration issue in the benchmark harness under
which some reasoning-model responses were cut off before a visible
answer was produced and scored as errors. With the harness corrected,
GPT-5.3's actual accuracy is 64\%. The correction is detailed in Section
6.8.

GPT-5.4 was reported in March 2026 to fail on the pioneer override
boundary: \texttt{access\_500m\_design} was incorrectly rejected (the
override applies at ≥ £500 M), while the identical scenario at £800 M
was correctly accepted. \textbf{In April 2026 this specific cell no
longer reproduces:} GPT-5.4 returns the correct verdict (covered) on
\texttt{access\_500m\_design} under the same prompt. GPT-4.1-mini's
enhanced-cover-chain failure is still present in v3.8 replication. The
engine evaluates \texttt{500\ \textgreater{}=\ 500\ =\ True} with no
ambiguity, regardless of model drift.

\subsubsection{6.4.1 v3.8 Adversarial Extension (20
scenarios)}\label{v3.8-adversarial-extension-20-scenarios}

The v3.7 58-scenario suite no longer reliably differentiates the
strongest frontier configurations from the Eligibility Module --- the
specific failure cells documented in March 2026 have largely closed at
current model performance. To test whether frontier models still fail at
deeper composition on the same domain, and whether the advantage of
deterministic execution is durable or transient, we authored 20 new
construction-CAR scenarios stratified across five complexity dimensions:

\begin{itemize}
\tightlist
\item
  \textbf{A --- Maximum-stack adversarial:} three or more rule-failure
  modes simultaneously (e.g.~depth-6 pioneer override + non-JCT
  existing-structures + design defect + access damage).
\item
  \textbf{B --- Threshold-boundary edge cases:} at-threshold (£100 M,
  £500 M), just-below (£99 M, £499 M), just-above (£100 M+ε, £500 M+ε).
\item
  \textbf{C --- Multi-clause cross-reference:} existing-structures + JCT
  + design + pioneer interactions across Cl. 5, 9, 10.
\item
  \textbf{D --- Surface-vs-deep contradiction:} surface text suggests
  opposite of correct answer (e.g.~``£800 M project, design defect,
  access damage'' surface signal = not covered; correct answer = covered
  via pioneer override).
\item
  \textbf{E --- Conjunction tracking:} every condition near-violating;
  one boolean flip reverses the outcome.
\end{itemize}

By design, these 20 scenarios stress-test specific compositional failure
modes. They are not claimed to be representative of typical legislative
drafting frequency or distribution; they are an adversarial probe at the
upper end of the complexity stack the Module is intended to handle,
included to test whether the engine's advantage persists when frontier
models are pushed beyond the v3.7 baseline difficulty.

\textbf{Methodology --- addressing the code-derived ground-truth
critique.} Each scenario was authored with an \emph{independent prose
justification} for its expected outcome before the engine was consulted.
The prose lives in the scenario's \texttt{notes:} field; the binary
expected-value (\texttt{expect.outcome}) was authored in the same pass
for consistency. The scenarios were then submitted to the engine via the
public \texttt{/api/v1/public/decide} endpoint against the canonical
bundle \texttt{construction-all-risks:20260412-gold} (the same bundle
the full 74-scenario replication suite of Table 8a uses; the rule
encoding was \emph{not} modified for the v3.8 extension). Engine outcome
matched authored expectation on 20/20, indicating self-consistency
between the prose, the binary, and the rule encoding. The frontier-LLM
runs (below) provide a third independent check: when models read the
same \texttt{source.md} and answer the same scenarios, their
disagreements with engine reveal model failures, not authoring errors
(see §6.4.1 cross-evaluation note).

\textbf{Frontier-LLM evaluation.} The 20 new scenarios were run against
four frontier-LLM configurations using the paper's exact
\texttt{\_build\_prompt} format from
\texttt{confidently-wrong-benchmark/benchmarks/run\_llm\_comparison.py}.
All five evaluations (engine + 4 LLM configs) ran independently against
the same scenario inputs.

\textbf{Table 8c: v3.8 Adversarial Extension --- 20 scenarios, 4
frontier-LLM configurations}

{\def\LTcaptype{none} % do not increment counter
\begin{longtable}[]{@{}
  >{\raggedright\arraybackslash}p{(\linewidth - 6\tabcolsep) * \real{0.3804}}
  >{\centering\arraybackslash}p{(\linewidth - 6\tabcolsep) * \real{0.1522}}
  >{\centering\arraybackslash}p{(\linewidth - 6\tabcolsep) * \real{0.1630}}
  >{\raggedright\arraybackslash}p{(\linewidth - 6\tabcolsep) * \real{0.3043}}@{}}
\toprule\noalign{}
\begin{minipage}[b]{\linewidth}\raggedright
Configuration
\end{minipage} & \begin{minipage}[b]{\linewidth}\centering
Correct
\end{minipage} & \begin{minipage}[b]{\linewidth}\centering
Wilson 95\% CI
\end{minipage} & \begin{minipage}[b]{\linewidth}\raggedright
Failure scenarios
\end{minipage} \\
\midrule\noalign{}
\endhead
\bottomrule\noalign{}
\endlastfoot
\textbf{Eligibility Module} & \textbf{20/20 (100\%)} &
\textbf{{[}83.9--100{]}} & --- \\
GPT-5.4 (low reasoning effort) & 20/20 (100\%) & {[}83.9--100{]} &
--- \\
Claude Sonnet 4.6 & 19/20 (95.0\%) & {[}76.4--99.1{]} & E4 (carveback
gap) \\
GPT-5.4 (default)* & 19/20 (95.0\%) & {[}76.4--99.1{]} & E4 (carveback
gap) \\
Claude Opus 4.7† & 18/20 (90.0\%) & {[}69.9--97.2{]} & B3 (£499 M
boundary), E4 \\
\end{longtable}
}

\emph{*GPT-5.4 (default) engaged no extended reasoning on any scenario;
see the reasoning-effort observation below. †Anthropic's strongest model
at evaluation time (April 2026). Scenario labels map to the released
YAML: E4 = \texttt{v38\_e4\_carveback\_gap\_explicit}, B3 =
\texttt{v38\_b3\_499m\_workmanship\_access\_no\_design\_limit}.}

\textbf{Figure 9: v3.8 adversarial extension forest plot.}

\pandocbounded{\includegraphics[keepaspectratio,alt={Wilson 95\% CIs across the five evaluated configurations on the 20-scenario v3.8 adversarial CAR extension. The Eligibility Module and GPT-5.4 at low reasoning effort sit at the 100\% ceiling; the other three frontier-LLM configurations each fail at least one scenario, with Opus 4.7 (Anthropic's strongest model at evaluation time, April 2026) failing two. Generated from Table 8c data by figures/scripts/forest\_plot\_v3\_8\_adversarial.py.}]{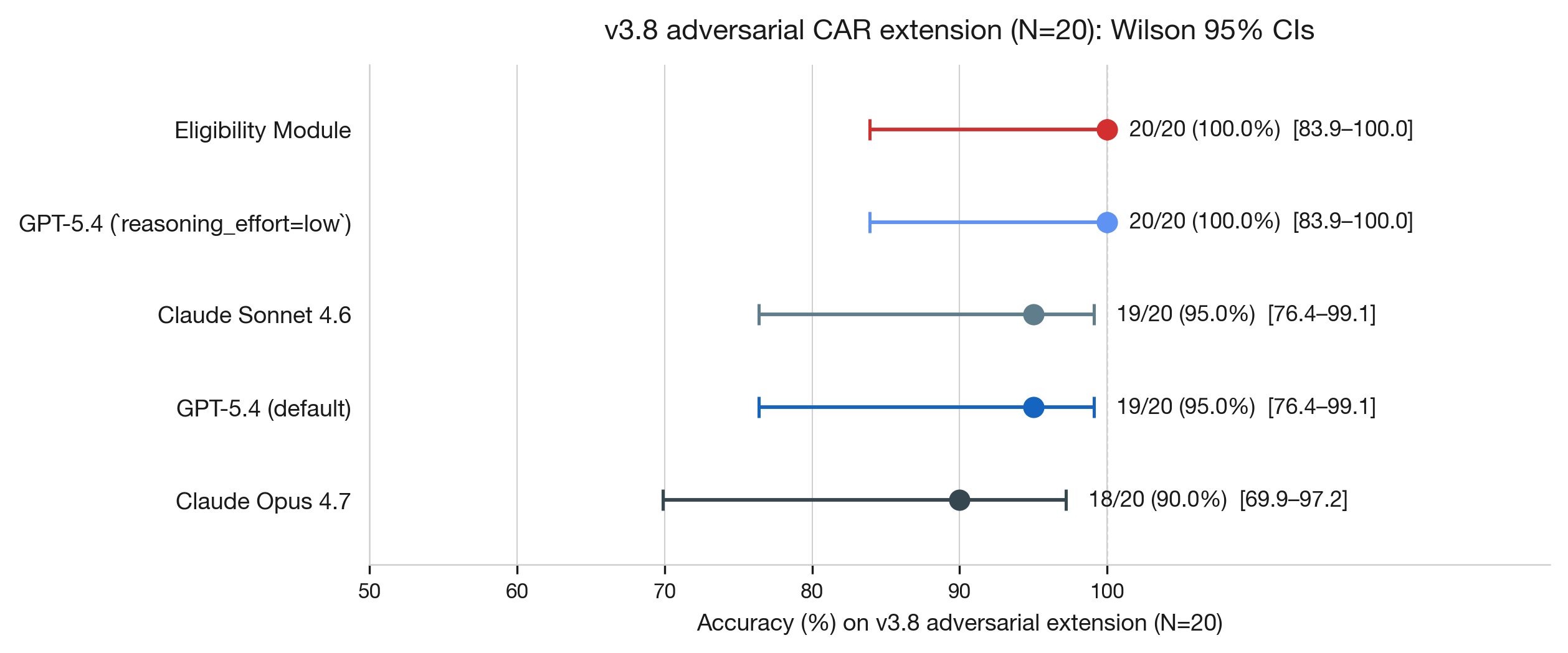}}

\textbf{Cross-evaluation note: the carveback-gap scenario (E4) is failed
by three of four frontier-LLM configurations across both Anthropic and
OpenAI families.} The scenario tests the DE3/LEG3 coverage gap
explicitly documented in \texttt{source.md} Cl. 7 commentary: with
\texttt{is\_access\_damage=true} and
\texttt{consequence\_of\_failure=false}, the carveback group fails
(Route A needs consequence; Route B needs not-access). The £200 M
project value qualifies for enhanced cover, but the prior carveback gate
blocks the claim before the enhanced-cover logic is reached. GPT-5.4
default and Sonnet 4.6 return \texttt{eligible}; Opus 4.7 returns
\texttt{eligible}; correct answer is \texttt{not\_eligible}. Only
GPT-5.4 at low reasoning effort --- the configuration that forces the
model to deliberate on the scenario rather than answer immediately (at
default effort it engaged no extended reasoning on any scenario) ---
gets it right. The pattern is consistent with a structural
compositional-evaluation failure mode rather than a per-model artefact.

\textbf{Reasoning-effort observation.} GPT-5.4 at default reasoning
effort engaged \textbf{no extended reasoning on any of the 20 v3.8
adversarial scenarios} --- it returns an immediate short answer with no
visible deliberation. At low reasoning effort, the same model
deliberates on every scenario. On the harder cases this additional
deliberation matters: low-reasoning gets E4 right where default
short-circuits to wrong. The v3.6 / v3.7 paper claim that ``default
reasoning is better than low reasoning'' (Finding 5, withdrawn) appears
to be inverted under current model behaviour --- default short-circuits
without deliberating; low forces deliberation.

Replication artefacts: \texttt{tools/replication\_run.py},
\texttt{tools/run\_engine\_v3\_8\_adversarial.py}, scenario YAML at
\texttt{dataset/construction-all-risks/scenarios\_v3\_8\_adversarial.yaml},
per-scenario JSONs at \texttt{docs/replication/B3\_*.json} and
\texttt{docs/replication/B4\_*.json}, full report at
\texttt{docs/replication/STREAM\_B\_REPORT.md}. All in the public
\texttt{confidently-wrong-benchmark/legalbench/} (paths relative to that
repo root).

\subsection{6.5 Error Analysis}\label{error-analysis}

Six findings emerge from the multi-model benchmark (one, Finding 5, was
withdrawn in v3.8 and is retained below for the record):

\textbf{Finding 1: The failure pattern is exception-chain specific.} All
frontier models evaluated on \texttt{english\_language} achieve 100\%
(43 scenarios with multi-route logic; §6.1 coverage matrix). On the
spacecraft section with its three-level exception chain, Claude Opus 4.6
drops to 90\% and GPT-5-mini to 75\%. The breaking point is nested
exceptions, not legal reasoning in general.

\textbf{Finding 2: The baseline failures are false negatives.} Across
the March 2026 baseline evaluations (§6.3--§6.4), we observed no false
positives: no LLM ever declared an ineligible applicant eligible or a
non-covered claim covered. The baseline failure mode is exclusively
conservative - but in a high-stakes context, a false negative may deny
someone a pathway the rules explicitly provide. Two later evaluation
arms are explicit exceptions to the no-false-positive observation: the
§6.4.1 v3.8 adversarial extension, where three frontier configurations
(GPT-5.4 default, Sonnet 4.6, and Opus 4.7) return \texttt{eligible} on
the \texttt{not\_eligible} E4 carveback-gap scenario --- false positives
(§6.4.1); and the §6.10 LegalBench classification baselines, which
exhibit positive-class anchoring under the upstream few-shot prompt
format, also producing false positives (see §6.10.4).

\textbf{Finding 3: The veteran exemption is the universal failure
point.} The most-failed scenario across all models is
\texttt{age\_59\_veteran\_1000hrs}. Claude Opus 4.6 and Sonnet 4.6 mark
this person ineligible on every run (0/3); GPT-5-mini answers correctly
on only 1 of 3 runs; GPT-5.4 alone is correct on all runs (Table 2). The
LLM conflates two independent exemption pathways.

\textbf{Finding 4 (revised v3.8): Frontier model accuracy is unstable
across model updates and prompt configurations.} The v3.6 / v3.7 paper
reported that GPT-5.4 dropped to 96.6\% on the construction insurance
section while achieving 100\% on spacecraft and immigration. \textbf{The
v3.8 replication finds that specific failure case has closed:} under the
same paper-prompt format and harness, current \texttt{gpt-5.4} is 74/74
(100\%) on the full construction-CAR suite (Table 8a). The Opus 4.6
spacecraft accuracy also rose from 89.7\% (March) to 98.5\% (April)
under the same \texttt{claude-opus-4-6} alias. No model-version bump was
announced; the specific empirical failures changed silently between
months. The structural failure pattern remains observable on the v3.8
adversarial extension (§6.4.1: Opus 4.7 fails 2/20, GPT-5.4 default
fails 1/20, Sonnet 4.6 fails 1/20), and on §6.10 LegalBench (combined
McNemar's \emph{p} ≤ 0.003 vs each frontier model; Table 6.10.B). For a
regulated workflow built on a benchmark-time accuracy claim, the
practical consequence is that specific model performance is a moving
target --- even within a single model alias --- while the deterministic
execution layer does not move.

\textbf{Finding 5 --- WITHDRAWN in v3.8.} Earlier revisions of this
paper (v3.6, v3.7) reported, as a tentative N=11 finding, that GPT-5.4
at low reasoning effort scored 7/11 (63.6\%) on the construction-CAR
exception-chain subset, matching GPT-5.3 at default reasoning.
\textbf{The v3.8 polish pass could not reproduce this result.} Two
converging issues: (a) the committed LLM-comparison harness in
\texttt{confidently-wrong-benchmark/benchmarks/run\_llm\_comparison.py}
does not exercise the reduced reasoning-effort configuration, and no
other committed script produced the v3.6 / v3.7 figure --- the original
result has no committed-code provenance; (b) an instrumented v3.8
replication on the exact 11 \texttt{exception\_chain}-tagged scenarios,
with the low reasoning-effort configuration explicitly requested and
full per-call logging, returns 11/11 (100\%) correct, with every
response complete and non-empty and with deliberate reasoning activity
recorded on each call. The parsimonious explanation is that the v3.6 /
v3.7 result was a harness artefact analogous to the v3.5 → v3.6 GPT-5.3
harness-configuration bug (§6.8 \emph{Harness configuration
correction}); we cannot rule out model drift since the original test,
but the absence of a reproducible script is on its own sufficient to
retract. Full withdrawal note, replication command, and per-scenario
JSON: \texttt{docs/r5-withdrawal-note.md} and
\texttt{docs/verify\_gpt5\_reasoning\_n11.json}. Section 6.9's
pre-registered N=66 replication is retained as the venue in which any
larger-sample reasoning-effort effect would be detected.

\textbf{Finding 6 (revised v3.8): The shifting-ground problem.} Multiple
v3.6 / v3.7 frontier-LLM cells do not replicate under the same prompt
and harness six weeks later. The most concrete examples: GPT-5.4 on
construction-CAR moved from 96.6\% to 100\% (Table 8a); Opus 4.6 on
spacecraft moved from 89.7\% to 98.5\% (Table 6); the v3.6 / v3.7
GPT-5.4 reasoning-effort-dependence claim (Finding 5) was withdrawn
after an instrumented replication produced the opposite result; the
GPT-5.3 model alias was deprecated by OpenAI mid-paper-cycle. None of
these are catastrophic individually, but the pattern is that
\emph{frontier-LLM accuracy on a fixed benchmark is a function of the
model snapshot, the harness configuration, and the prompt format --- and
at least one of these can shift without notice}. For a regulated
workflow that depends on benchmark-time accuracy claims to certify
deployment, this property is structurally incompatible with the
verification pipelines required by frameworks like the EU AI Act. The
Eligibility Module avoids this class of risk: rules are compiled once at
authoring time and evaluated deterministically thereafter; the same
bundle gives the same answer on the same inputs in March, April, or any
subsequent month, regardless of any silent change to upstream model
behaviour. The v3.8 adversarial extension (§6.4.1) is published
precisely so that this paper's frontier-LLM numbers are themselves
replicable from the committed harness --- not as a snapshot but as a
procedure.

\subsection{6.6 The Cost-Accuracy
Trade-off}\label{the-cost-accuracy-trade-off}

{\def\LTcaptype{none} % do not increment counter
\begin{longtable}[]{@{}
  >{\raggedright\arraybackslash}p{(\linewidth - 4\tabcolsep) * \real{0.2041}}
  >{\centering\arraybackslash}p{(\linewidth - 4\tabcolsep) * \real{0.3878}}
  >{\centering\arraybackslash}p{(\linewidth - 4\tabcolsep) * \real{0.4082}}@{}}
\toprule\noalign{}
\begin{minipage}[b]{\linewidth}\raggedright
Property
\end{minipage} & \begin{minipage}[b]{\linewidth}\centering
Eligibility Module
\end{minipage} & \begin{minipage}[b]{\linewidth}\centering
GPT-5.4 (best LLM)
\end{minipage} \\
\midrule\noalign{}
\endhead
\bottomrule\noalign{}
\endlastfoot
Deterministic & Yes & No \\
Cost per evaluation & Near-zero marginal after compilation &
\textasciitilde\$0.02 per scenario \\
Latency & \textless1ms & \textasciitilde2--5 seconds \\
Auditability & Exact logical path & Post-hoc rationalisation \\
Regulatory defensibility & Execution semantics defined & Statistical
accuracy \\
Model deprecation risk & None & Version-dependent \\
\end{longtable}
}

A production system cannot guarantee that GPT-5.4's 100\% accuracy on
this benchmark extends to all possible inputs. It is an empirical
observation, not an execution-level guarantee.

\subsection{6.7 Robustness Analysis}\label{robustness-analysis}

This section addresses the key question: whether the observed failures
are prompt-sensitive or structurally robust. The central question is not
``what is the best possible LLM score after bespoke optimisation?'' but
``when models fail on exception chains, is that a prompting artefact -
fixable with better instructions - or a deeper reliability issue?''

\textbf{Table 9: Robustness Analysis (spacecraft section, 68 scenarios)}

{\def\LTcaptype{none} % do not increment counter
\begin{longtable}[]{@{}
  >{\raggedright\arraybackslash}p{(\linewidth - 4\tabcolsep) * \real{0.6119}}
  >{\centering\arraybackslash}p{(\linewidth - 4\tabcolsep) * \real{0.2836}}
  >{\centering\arraybackslash}p{(\linewidth - 4\tabcolsep) * \real{0.1045}}@{}}
\toprule\noalign{}
\begin{minipage}[b]{\linewidth}\raggedright
Condition
\end{minipage} & \begin{minipage}[b]{\linewidth}\centering
Claude Opus 4.6
\end{minipage} & \begin{minipage}[b]{\linewidth}\centering
GPT-5.4
\end{minipage} \\
\midrule\noalign{}
\endhead
\bottomrule\noalign{}
\endlastfoot
Generic prompt, T=0.3 (Claude) / API default (GPT), N=3
\emph{(temperature-sensitivity arm)} & 90\% (7 FN) & 100\% \\
Generic prompt, T=0, N=3 \emph{(primary configuration; Claude only)} &
90\% (7 FN, all 0/3) & N/A* \\
Generic prompt, T=0.3, N=10 \emph{(Claude only)} & 90\% (7 FN, all
\textbf{0/10}) & - \\
Enhanced prompt, N=3 & \textbf{65\%} (4 FN + \textbf{20 FP}) &
\textbf{99\%} (1 FN) \\
\end{longtable}
}

\emph{*GPT-5 family does not expose a temperature parameter.}

\textbf{R1 (March 2026 snapshot; partial v3.8 caveat).} At temperature 0
(fully deterministic), Claude Opus produced the same 7 failures with the
same 0/3 error rate at the time of the original test. With 10 runs per
scenario, all 7 failed scenarios scored \textbf{0/10}. Aggregated across
the 7 scenarios, this is \textbf{0 correct answers in 70 independent
trials}. The Clopper--Pearson 95\% one-sided upper bound on per-trial
success probability is \textbf{4.19\%}, ruling out the hypothesis that
the model would produce correct answers at a rate ≥5\% per attempt at
that snapshot.

\textbf{v3.8 caveat.} The ``7 failing spacecraft scenarios'' baseline
that R1 was tested on (Opus 4.6 = 61/68 in March) does not replicate
today: in April 2026 Opus 4.6 returns the correct answer on 67 of 68
spacecraft scenarios under the same prompt and harness, leaving only 1
systematic failure rather than 7 (Table 6 v3.8 column). The 70-trial
0/70 result therefore stands as a March 2026 finding about a specific
7-scenario set that is no longer at the same baseline failure rate.
Re-running the 70-trial design against the current 1-scenario failure
set would require a separate investigation we have not undertaken in
v3.8. The headline qualitative claim --- \emph{that some specific Opus
failures were stable rather than stochastic at the original snapshot}
--- survives. The headline quantitative claim --- \emph{that 4.19\%
upper bound applies to current Opus 4.6 production behaviour on current
failure cases} --- does not.

\textbf{R2: Enhanced prompting trades one failure mode for another.} The
enhanced prompt explicitly instructs the LLM to ``evaluate each
exemption independently'' and warns about exception chains. This
partially succeeds: the false-negative count falls from 7 to 4 (with
run-to-run variance in which specific scenarios fail; see R4). However,
the same instructions cause the model to over-apply exemption logic,
producing \textbf{20 false positives} --- the first time in our
benchmark that an LLM declares an ineligible applicant eligible. The
model now grants eligibility to applicants below the flight hours
threshold (499 \textless{} 500), with expired medical certificates, and
on orbital missions where the age exemption is explicitly revoked. Net
accuracy drops from 61/68 (89.7\%, CI {[}80.2\%, 94.9\%{]}) to 44/68
(64.7\%, CI {[}52.8\%, 75.0\%{]}) --- the CIs are disjoint, so the drop
is statistically significant.

\textbf{Figure 5: Prompt-repair trade-off (Claude Opus 4.6,
spacecraft).}

\pandocbounded{\includegraphics[keepaspectratio,alt={Accuracy with Wilson 95\% CIs under the generic and enhanced prompt, with per-row outcome composition annotated inline (correct / false-negative / false-positive counts out of n=68). The 89.7\% → 64.7\% drop is statistically significant --- the two CIs are disjoint, shaded in red. The generic prompt fails conservatively (7 FN, 0 FP); the enhanced prompt reduces the false-negative count from 7 to 4 but introduces 20 false positives for the first time in the benchmark. Generated from Section 6.7 data by figures/scripts/fnfp\_tradeoff\_r2.py.}]{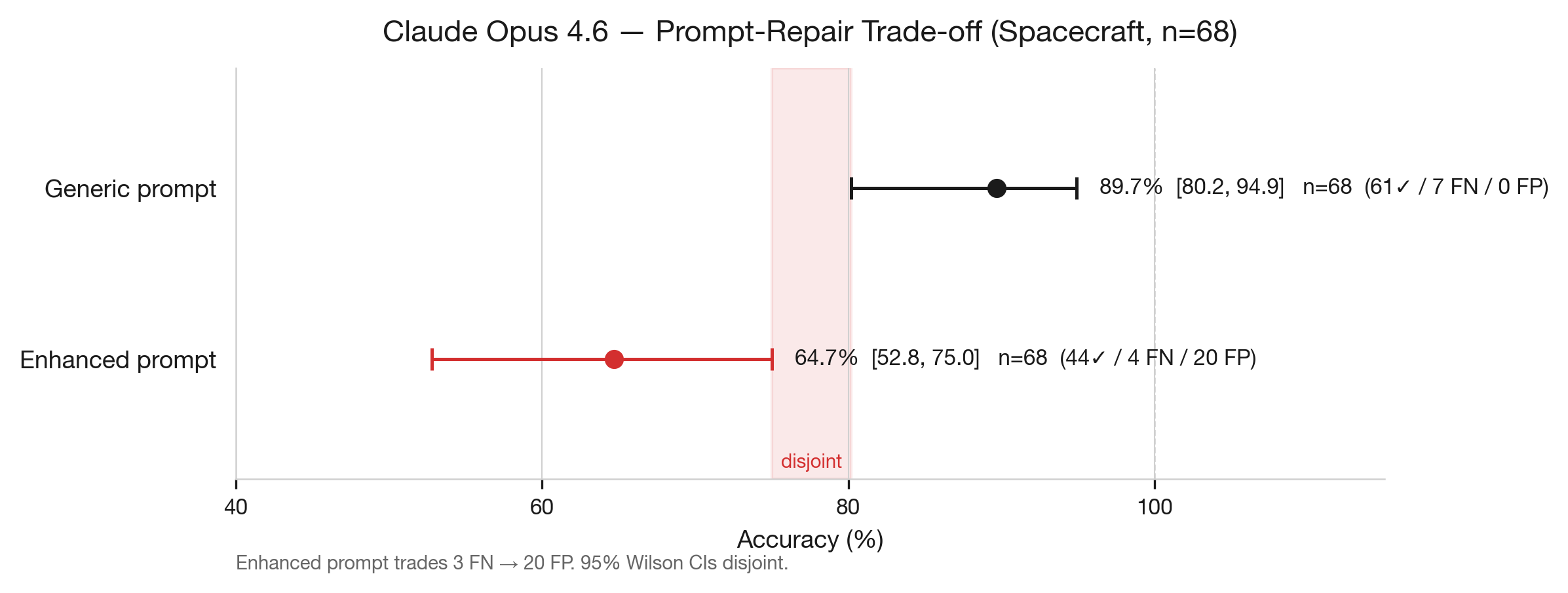}}
We note that R2 tests only one enhanced prompt variant. Section 6.9
pre-registers a three-variant comparison (few-shot, self-critique,
decomposition) to test whether this FN/FP trade-off is fundamental to
prompt repair on this class of task or specific to the particular prompt
tested here.

This is not simply a bad prompt. The enhanced prompt correctly
identifies the reasoning gap (independent exemption evaluation) and
provides accurate instructions. The problem is that the same
instructions that fix under-application of exemptions cause
over-application elsewhere. The generic prompt's conservative bias - all
false negatives, no false positives - is a more predictable failure mode
than the enhanced prompt's scattered errors across both directions. For
a production system, a consistent conservative bias is easier to
compensate for than unpredictable failures in both directions.

\textbf{R3: GPT-5.4 shows the same pattern in miniature.} The enhanced
prompt introduces 1 failure on the hardest scenario that the generic
prompt handles correctly.

\textbf{R4: Run-to-run variance confirms fragility.} Re-running the
enhanced prompt benchmark on the same model produces qualitatively
identical results (20 FP, 4 FN) but with slight variance in which
specific scenarios fail. This contrasts with the generic prompt, where
the same 7 scenarios fail identically across independent runs. The
enhanced prompt introduces not only new failure modes but also
non-deterministic failure patterns - precisely the property a
high-stakes system cannot tolerate.

\textbf{R5 --- WITHDRAWN in v3.8.} Earlier revisions of this paper
reported that GPT-5.4 at low reasoning effort scored 7/11 on the
construction-CAR exception chain, suggesting
reasoning-compute-dependence within the GPT-5 family. The v3.8 polish
pass found (a) no committed script reproduces the v3.6 / v3.7 result ---
the LLM-comparison harness at
\texttt{confidently-wrong-benchmark/benchmarks/run\_llm\_comparison.py}
does not exercise the reduced reasoning-effort configuration, and (b) an
instrumented replication on the exact 11
\texttt{exception\_chain}-tagged scenarios, with the low
reasoning-effort configuration explicitly requested, returns 11/11
(100\%) correct, with every response complete and non-empty and
deliberate reasoning activity recorded on each call. Most likely cause:
the v3.6 / v3.7 figure was a harness artefact analogous to the v3.5 →
v3.6 GPT-5.3 harness-configuration bug. The intra-model
reasoning-compute-dependence hypothesis is therefore \textbf{withdrawn}.
Claude Opus 4.6's temperature stability (identical spacecraft results at
T=0 and T=0.3; Table 9) stands as a separate observation but no longer
serves as a parallel data point to a withdrawn claim. Section 6.9's
pre-registered N=66 replication is retained as the venue for any
larger-sample test of reasoning-effort dependence. Withdrawal note +
per-scenario JSON: \texttt{docs/r5-withdrawal-note.md},
\texttt{docs/verify\_gpt5\_reasoning\_n11.json}.

The robustness findings address the prompting-contest objection at the
level of the strategies tested here. The gap is not caused by a
suboptimal generic prompt that the enhanced variant fixes. On the two
prompts tested, prompt engineering on this class of task exhibits a
trade-off: instructions that reduce false negatives on exception chains
simultaneously increase false positives by causing over-application of
the same exemption logic. This is consistent with the compositional
reasoning limitation identified by Dziri et al.
\citeyearpar{dziri2023faith}: the model does not have a stable internal
representation of the exception chain structure that can be steered by
prompt instructions without side effects. We find no evidence that the
prompting strategies tested here reliably resolve the failure mode;
whether richer strategies (few-shot, self-critique, decomposition,
tool-augmented) close the gap is the pre-registered subject of §6.9.

\subsection{6.8 Threats to Validity}\label{threats-to-validity}

We follow academic convention in identifying threats to the validity of
these findings.

\textbf{Internal validity:} - \emph{Prompt sensitivity.} The benchmark
uses a generic prompt, not an optimised one. This is intentional
(Section 6.1), but means we are not measuring best-possible LLM
performance. The robustness analysis (Section 6.7) tests one enhanced
prompt variant that specifically targets the identified failure pattern;
it reduces false negatives but introduces false positives, with net
accuracy decreasing. We cannot exclude the possibility that a different
prompt strategy could improve accuracy without side effects, but the
result demonstrates that targeted prompt repair on this class of task
involves trade-offs between failure modes. - \emph{Harness configuration
correction (v3.6).} v3.5 reported GPT-5.3 at 27\% (3/11) on the
construction exception chain. That figure was inflated by a
configuration issue in the benchmark harness under which some
reasoning-model responses were cut off before a visible answer was
produced, and the resulting empty responses were scored as errors. With
the harness corrected, GPT-5.3's actual accuracy is 64\% (7/11). The
correction does not affect the qualitative finding --- GPT-5.3 still
fails systematically on multi-level exception scenarios --- but the
quantitative severity is less extreme than originally reported. The
benchmark script has been corrected and the updated results are publicly
reproducible. - \emph{Expected values partly code-derived.} The
\texttt{life\_uk} expected values are computed by Python code from the
section's authored eligibility formula (Appendix A.1); L.D. authored and
validated the legal formalisation (Author Contributions), but the 56
generated expected values were not individually hand-verified. The
\texttt{english\_language} scenarios were hand-verified against Form AN
guidance. The spacecraft scenarios use a synthetic statute with
unambiguous values. - \emph{Structured inputs only.} Applicant data is
provided as typed key-value pairs. A production system requires
extraction from natural language or documents; that layer is not tested
here. - \emph{Adversarial scenarios are post-hoc.} The spacecraft
adversarial suite was designed after observing LLM failure patterns. The
baseline results (48 scenarios) show the same failure pattern; the
adversarial additions amplify it. Selection bias of this kind is
addressed for the LegalBench external-validation arm by §6.10.3's
pre-registered random-sample replication (seeds 42 and 43 run; seed 44
pre-registered but not yet run).

\textbf{External validity:} - \emph{Synthetic domains.} The spacecraft
domain uses a synthetic statute. The construction insurance domain uses
synthetic policy wording modelled on real clauses. Findings in these
domains may not generalise directly to all legal contexts. §6.10
provides cross-domain external validity on a public peer-reviewed
legal-reasoning benchmark (LegalBench). - \emph{Benchmark scope.} The
benchmark tests one specific class of task --- nested exception-chain
evaluation. LLM performance on other legal reasoning tasks may differ
substantially; §6.10 reports both the multi-prong
(failure-pattern-matching) and single-clause-classification
(random-sample) regimes. - \emph{LegalBench prompt format.} The low
GPT-5.4 cells in §6.10.4 are tied to the upstream LegalBench few-shot
classification prompt format. We report them because they affect the
canonical baseline under the evaluated protocol, but they should not be
read as best-achievable GPT-5.4 performance estimates. A broader prompt
sweep would be required before treating those cells as model-capability
estimates. - \emph{Model versioning.} Specific model versions are named.
As models are updated, results may change. The benchmark scenarios are
released publicly to allow ongoing evaluation of future systems. The
specific accuracy figures reported here are a snapshot; the benchmark
task structure and failure pattern are the durable contribution.

\textbf{Construct validity:} - \emph{LLM self-evaluation in quality
engineering.} Quality scores in Section 7 are produced by an LLM
evaluator, not human legal experts. These are internal consistency
scores; external validation has not been performed.

\subsection{6.9 Pre-Registered Replication
(v3.7)}\label{pre-registered-replication-v3.7}

The prompt-repair finding (R2) rests on one enhanced prompt variant; the
reasoning-effort question (formerly Finding 5 / R5, withdrawn in v3.8
--- see §6.5) is now an open hypothesis rather than a tentative finding.
v3.7 of this paper pre-registers the following replication, which v3.8
retains as the venue for both questions:

\begin{enumerate}
\def\labelenumi{\arabic{enumi}.}
\item
  \textbf{Expanded construction\_car exception-chain subset from 11 to
  66 scenarios}, stratified across five depth bands (simple exclusion,
  exclusion+carve-back, +re-exclusion, +enhanced cover,
  +design-limit/pioneer/depth-5) with boundary and adversarial coverage
  within each band. Power analysis: N=35 per arm is required to detect a
  64\% vs 91\% effect at 80\% power, α=0.05; N=66 gives
  \textasciitilde1.9× margin and tightens the Wilson 95\% CI at 64\%
  observed accuracy from ±25pp (at N=11) to ±11pp.
\item
  \textbf{Three prompt-variant robustness tests} on the 7 known-failing
  spacecraft scenarios plus 3 control scenarios (N=3 runs each), across
  Claude Opus 4.6 and GPT-5.4: (a) few-shot with worked exception-chain
  examples, (b) self-critique two-stage, (c) decomposition (evaluate
  each route separately, then OR). Primary outcome: does any variant
  escape the FN/FP trade-off observed in R2?
\item
  \textbf{Full reasoning-effort sweep} on the 66-scenario exception
  chain: GPT-5.4 at low/medium/high, GPT-5.3 at default (and at low/high
  if the parameter is exposed), Claude Opus 4.6 as invariance control.
\item
  \textbf{Independent expected-value authoring for new scenarios}: each
  of the 55 new scenarios has a prose expected-value assertion written
  against the legislative clause chain before the fixture is consulted.
  Mismatches between prose and fixture are flagged for SME review rather
  than silently reconciled. This addresses the internal-validity hole
  where code-derived expected values allow fixtures and the execution
  engine to trivially agree.
\end{enumerate}

\textbf{Pre-registration commitment.} Results --- whether they confirm,
weaken, or refute the earlier findings --- will be reported in a
subsequent revision regardless of direction. (Status as of v3.11, July
2026: 16 of the 55 pre-registered new exception-chain scenarios were
authored and exercised in the v3.8 construction-CAR replication --- they
form the 58→74 expansion behind Table 8a --- but the full N=66
sub-study, with its three prompt-variant arms and reasoning-effort
sweep, remains pre-registered and not yet run; it is the next
experimental milestone for this paper.) Findings 5 and 6, and Section
6.7 R5, will be updated or withdrawn based on the replication.

\begin{center}\rule{0.5\linewidth}{0.5pt}\end{center}

\subsection{6.10 External Validation on
LegalBench}\label{external-validation-on-legalbench}

The benchmark in §6.1--§6.8 isolates a single failure pattern (nested
exception-chain evaluation) on synthetic and UK-immigration scenarios
authored for the experiment. A natural external test of the architecture
is a public, peer-reviewed legal-reasoning benchmark covering tasks the
authors did not design. We take that test using LegalBench
\citeyearpar{guha2023legalbench}, evaluating the Eligibility Module on
nine LegalBench tasks (949 held-out cases) against three frontier LLMs
spanning two model families.

\subsubsection{6.10.1 Setup}\label{setup}

For each LegalBench task we evaluate, we author one rule bundle for the
Eligibility Module from the \textbf{verbatim canonical Task description}
published in the upstream LegalBench repository (CC BY 4.0, Neel Guha).
No paraphrase, no editorial shaping, no DSL pseudo-code. The bundle is
generated by aethis-core's authoring pipeline against a small set of
statute-derived test cases written by the authors (typically 4--8 per
task) --- \emph{not} drawn from the LegalBench fact patterns. Where
domain-specific subject-matter-expert guidance is added, it lives in a
separate \texttt{guidance/} directory and is grounded in independent
practitioner authority (e.g.~LSTA Model Credit Agreement conventions for
\texttt{cuad\_covenant\_not\_to\_sue}; standard federal civil-procedure
treatments for \texttt{personal\_jurisdiction}; the Atticus Project's
CUAD taxonomy for clause-classification tasks).

We compare the \textbf{Eligibility Module pipeline} (LLM extractor
producing typed atomic field values \(\to\) deterministic engine
evaluation against the bundle's compiled rule) against the
\textbf{LLM-only baseline} (the same model given the canonical rule
prose and asked to answer the LegalBench question end-to-end, using
upstream LegalBench prompts --- the canonical 5--6-shot
\texttt{base\_prompt.txt} files). LLM baselines are evaluated at the
model level for \texttt{claude-sonnet-4-6}, \texttt{claude-opus-4-7},
and \texttt{gpt-5.4}. The Eligibility Module's runtime extractor in the
headline configuration is \texttt{claude-sonnet-4-6}; we also report a
cross-extractor configuration with \texttt{claude-opus-4-7} to
disentangle structural from model-specific effects (§6.10.5).

For tasks selected by inspection (matching the failure pattern
identified in §4.2), we apply the held-out methodology
\emph{retroactively}: results below are filtered to the held-out half of
each task's LegalBench \texttt{test} split using a deterministic seeded
partition
(\texttt{tools/test\_split.py\ -\/-seed\ 7\ -\/-dev\_fraction\ 0.5}).
For tasks selected by \textbf{pre-registered random sample} from the
153-task LegalBench candidate pool --- the 162 upstream tasks minus the
nine already integrated at pick time
(\texttt{tools/random\_task\_pick.py\ -\/-seed\ 42} and
\texttt{-\/-seed\ 43}, committed before authoring), the held-out
methodology was applied \emph{prospectively} --- hint iteration was
confined to the dev half, and the holdout was evaluated once with frozen
hints.

\subsubsection{6.10.2 Results}\label{results-1}

Table 6.10.A reports per-task accuracy on the held-out evaluation
sample, with 95\% Wilson confidence intervals on each estimate and exact
two-sided McNemar's tests on the per-case engine-vs-LLM agreement
matrices.

\textbf{Table 6.10.A --- Eligibility Module vs frontier LLMs, LegalBench
held-out evaluation}

\small

{\def\LTcaptype{none} % do not increment counter
\begin{longtable}[]{@{}
  >{\raggedright\arraybackslash}p{(\linewidth - 10\tabcolsep) * \real{0.1579}}
  >{\raggedleft\arraybackslash}p{(\linewidth - 10\tabcolsep) * \real{0.2105}}
  >{\raggedright\arraybackslash}p{(\linewidth - 10\tabcolsep) * \real{0.1579}}
  >{\raggedright\arraybackslash}p{(\linewidth - 10\tabcolsep) * \real{0.1579}}
  >{\raggedright\arraybackslash}p{(\linewidth - 10\tabcolsep) * \real{0.1579}}
  >{\raggedright\arraybackslash}p{(\linewidth - 10\tabcolsep) * \real{0.1579}}@{}}
\toprule\noalign{}
\begin{minipage}[b]{\linewidth}\raggedright
Task
\end{minipage} & \begin{minipage}[b]{\linewidth}\raggedleft
N
\end{minipage} & \begin{minipage}[b]{\linewidth}\raggedright
Eligibility Module
\end{minipage} & \begin{minipage}[b]{\linewidth}\raggedright
Sonnet 4.6
\end{minipage} & \begin{minipage}[b]{\linewidth}\raggedright
Opus 4.7
\end{minipage} & \begin{minipage}[b]{\linewidth}\raggedright
GPT-5.4
\end{minipage} \\
\midrule\noalign{}
\endhead
\bottomrule\noalign{}
\endlastfoot
\texttt{hearsay} (FRE 801) & 47 & 45/47 (95.7\%) {[}85.8--98.8{]} &
35/47 (74.5\%); +21.3 / 0.006 & 39/47 (83.0\%); +12.8 / 0.031 & 40/47
(85.1\%); +10.6 / 0.125 \\
\texttt{personal\_jurisdiction} & 25 & 24/25 (96.0\%) {[}80.5--99.3{]} &
24/25 (96.0\%); +0.0 / 1.000 & 23/25 (92.0\%); +4.0 / 1.000 & 24/25
(96.0\%); +0.0 / 1.000 \\
\texttt{jcrew\_blocker} & 27 & 27/27 (100\%) {[}87.5--100{]} & 16/27
(59.3\%); +40.7 / \textless0.001 & 26/27 (96.3\%); +3.7 / 1.000 & 25/27
(92.6\%); +7.4 / 0.500 \\
\texttt{cuad\_covenant\_not\_to\_sue} ¹ & 154 & 150/154 (97.4\%)
{[}93.5--99.0{]} & 148/154 (96.1\%); +1.3 / 0.727 & 149/154 (96.8\%);
+0.6 / 1.000 & 83/154 (53.9\%); +43.5 / \textless0.001 \\
\texttt{contract\_nli\_explicit\_id} ¹ & 55 & 48/55 (87.3\%)
{[}76.0--93.7{]} & 47/55 (85.5\%); +1.8 / 1.000 & 46/55 (83.6\%); +3.6 /
0.688 & 20/55 (36.4\%); +50.9 / \textless0.001 \\
\texttt{contract\_nli\_notice\_on\_disclosure} ² & 71 & 70/71 (98.6\%)
{[}92.4--99.8{]} & 71/71 (100\%); −1.4 / 1.000 & 71/71 (100\%); −1.4 /
1.000 & 36/71 (50.7\%); +47.9 / \textless0.001 \\
\texttt{learned\_hands\_health} ² & 113 & 107/113 (94.7\%)
{[}88.9--97.5{]} & 104/113 (92.0\%); +2.7 / 0.375 & 99/113 (87.6\%);
+7.1 / 0.039 & 56/113 (49.6\%); +45.1 / \textless0.001 \\
\texttt{cuad\_liquidated\_damages} ² & 110 & 105/110 (95.5\%)
{[}89.8--98.0{]} & 107/110 (97.3\%); −1.8 / 0.500 & 105/110 (95.5\%);
+0.0 / 1.000 & 79/110 (71.8\%); +23.6 / \textless0.001 \\
\texttt{opp115\_int\_and\_specific\_audiences} ² & 347 & 324/347
(93.4\%) {[}90.3--95.5{]} & 319/347 (91.9\%); +1.4 / 0.302 & 321/347
(92.5\%); +0.9 / 0.581 & 232/347 (66.9\%); +26.5 / \textless0.001 \\
\textbf{All 9 tasks (combined)} & \textbf{949} & \textbf{900/949
(94.8\%)} & \emph{combined: see Table 6.10.B} & & \\
\end{longtable}
}

\normalsize

¹ Random sample, seed=42 (pre-registered in
\texttt{tools/random\_task\_pick.py}). Notice-on-compelled-disclosure
full task name:
\texttt{contract\_nli\_notice\_on\_compelled\_disclosure}. Explicit-id
full name: \texttt{contract\_nli\_explicit\_identification}. Opp115 full
name: \texttt{opp115\_international\_and\_specific\_audiences}.

² Random sample, seed=43.

\textbf{Cell legend.} Engine column: correct/n (accuracy, 95\% Wilson
CI). LLM columns: correct/n (accuracy); Δ vs Engine in percentage points
/ exact two-sided McNemar's \emph{p} on per-case discordance with the
Engine. Negative Δ indicates the LLM out-performed the Engine at that N;
no negative Δ reaches significance.

When per-task results are aggregated by paired-binomial test on
discordant cases across the full 949-case held-out evaluation, the
engine is significantly more accurate than each of the three frontier
models at the conventional \(p < 0.05\) level:

\textbf{Table 6.10.B --- Combined paired-binomial across 9 LegalBench
tasks}

{\def\LTcaptype{none} % do not increment counter
\begin{longtable}[]{@{}
  >{\raggedright\arraybackslash}p{(\linewidth - 6\tabcolsep) * \real{0.2000}}
  >{\raggedleft\arraybackslash}p{(\linewidth - 6\tabcolsep) * \real{0.2667}}
  >{\raggedleft\arraybackslash}p{(\linewidth - 6\tabcolsep) * \real{0.2667}}
  >{\raggedleft\arraybackslash}p{(\linewidth - 6\tabcolsep) * \real{0.2667}}@{}}
\toprule\noalign{}
\begin{minipage}[b]{\linewidth}\raggedright
Comparison
\end{minipage} & \begin{minipage}[b]{\linewidth}\raggedleft
\(b\) (Eligibility Module only correct)
\end{minipage} & \begin{minipage}[b]{\linewidth}\raggedleft
\(c\) (LLM only correct)
\end{minipage} & \begin{minipage}[b]{\linewidth}\raggedleft
Two-sided exact \emph{p}
\end{minipage} \\
\midrule\noalign{}
\endhead
\bottomrule\noalign{}
\endlastfoot
Module vs Sonnet 4.6 & 44 & 15 & \(< 0.001\) \\
Module vs Opus 4.7 & 34 & 13 & \(0.003\) \\
Module vs GPT-5.4 & 340 & 35 & \(< 0.001\) \\
\end{longtable}
}

The aggregate GPT-5.4 comparison should be interpreted cautiously. Most
of the discordant GPT-5.4 cases arise from the classification-prompt
sensitivity analysed in §6.10.4, not from the multi-prong
rule-evaluation tasks that provide the closest external analogue to the
exception-chain failure pattern in §4.2. We therefore treat the GPT-5.4
aggregate as evidence that the Module remains robust under the evaluated
LegalBench protocol, not as a broad claim about GPT-5.4's
best-achievable legal-reasoning performance.

\textbf{Per-task statistical power.} The small-N curated tasks have
limited per-task power: \texttt{personal\_jurisdiction} (N=25) cannot
detect held-out accuracy differences smaller than approximately ±20pp at
\(\alpha = 0.05\), 80\% power, and \texttt{hearsay} (N=47) is similarly
underpowered for \(\Delta < 14\)pp.~The primary inferential test in this
section is the combined paired-binomial across all 949 held-out cases
(Table 6.10.B); per-task McNemar's \emph{p}-values are reported for
completeness but should be interpreted as supplementary, particularly on
the curated multi-prong tasks.

\textbf{Multiple comparisons.} Table 6.10.A reports 27 simultaneous
per-task McNemar's tests (9 tasks × 3 LLMs), uncorrected. Under
Bonferroni correction at \(\alpha = 0.05/27 \approx 0.0019\), only the
strongest comparisons remain individually significant (most of the
GPT-5.4 cells; \texttt{jcrew\_blocker} vs Sonnet). The combined
paired-binomial in Table 6.10.B aggregates discordant pairs across all
tasks and is a single test, not subject to the same correction; it is
the primary inferential claim and remains \(p \le 0.003\) against every
model under any reasonable family-wise correction.

\textbf{Figure 7: LegalBench per-task held-out accuracy with Wilson 95\%
CIs.}

\pandocbounded{\includegraphics[keepaspectratio,alt={Forest plot of per-task held-out accuracy across the four evaluated configurations (Eligibility Module, Sonnet 4.6, Opus 4.7, GPT-5.4). The Eligibility Module is right-shifted on every task. GPT-5.4's positive-class anchoring on classification-style tasks (§6.10.4) is visible in the lower six rows where its CI sits well to the left of the other three. On the curated multi-prong tasks (top three rows), Sonnet 4.6's wider variability is visible (jcrew\_blocker 59.3\% vs the others' 92--96\%). Generated from Table 6.10.A data by figures/scripts/forest\_plot\_legalbench.py.}]{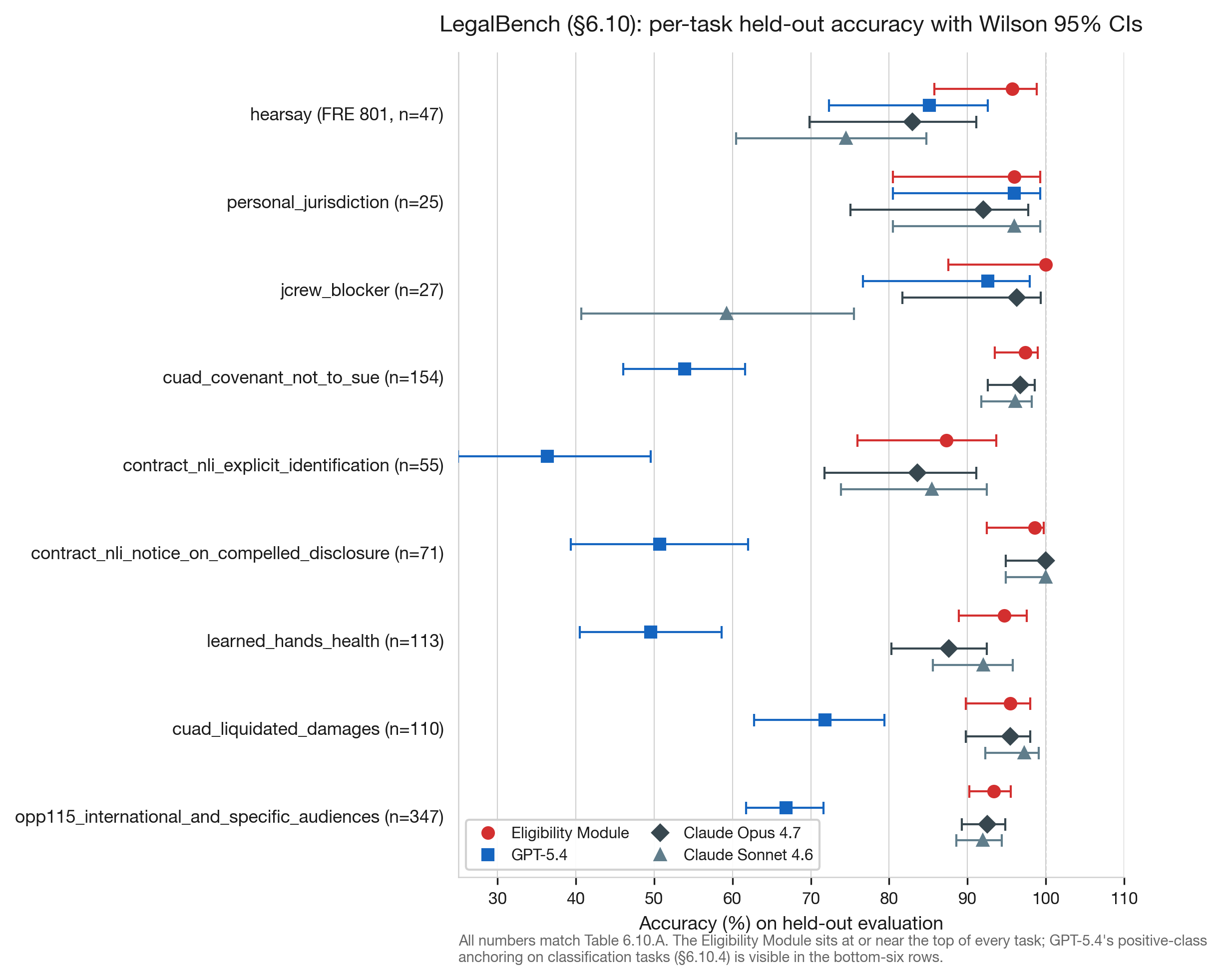}}

\subsubsection{6.10.3 Selection bias and the random-sample
replication}\label{selection-bias-and-the-random-sample-replication}

A reviewer might fairly ask whether the three ``curated'' tasks above
(\texttt{hearsay}, \texttt{personal\_jurisdiction},
\texttt{jcrew\_blocker}) were chosen because they fit the Eligibility
Module's structural shape --- multi-prong rule application of the kind
§4.2 identifies as a frontier-LLM failure pattern. They were. To test
whether the result is selection-bias-dependent, we repeated the protocol
on \textbf{six additional LegalBench tasks selected by seeded random
sample} from the 153-task candidate pool (the 162 upstream tasks minus
the nine already integrated at pick time). The selection (seeds 42 and
43, n=2 and n=4 respectively, generated by
\texttt{tools/random\_task\_pick.py}) was committed before the upstream
Task descriptions were inspected. A third seed (seed=44, n=2) was
pre-registered at the public release tag
\texttt{pre-v3.8-legalbench-preregistration} (annotated tag object
\texttt{f7c5994}; the tagged commit is \texttt{58d3b5d}) but is not yet
run; the tag preserves the seed-before-results discipline if those tasks
are added in a future revision.

The six random-sample tasks span CUAD contract-clause classification,
ContractNLI entailment, OPP115 privacy-policy classification, and
LearnedHands legal-aid post categorisation --- predominantly
single-clause classification rather than the multi-prong rule
application that §4.2 characterises as the failure pattern.
Predicted-mismatch territory.

The result is consistent: the Eligibility Module wins on the
random-sample tasks, but with smaller per-task margins against Sonnet
4.6 and Opus 4.7 (1--3 percentage points on the random sample, versus up
to +41 percentage points on the curated multi-prong tasks); GPT-5.4's
per-task margins are larger in both arms but are dominated by the
§6.10.4 prompt-format sensitivity rather than the exception-chain
pattern. The combined paired-binomial test reported in Table 6.10.B
includes both curated and random-sample tasks; with the curated subset
removed, the random-sample combined comparison reaches significance
against GPT-5.4 (\(b=331\), \(c=33\), \(p<0.001\)); against Opus 4.7 it
is suggestive but not significant (\(b=26\), \(c=13\), \(p=0.053\)), and
against Sonnet 4.6 it is positive but not significant (\(b=22\),
\(c=14\), \(p=0.24\)). (Correction, v3.11: earlier revisions printed
random-only aggregates computed under a mislabelled five-task partition
that counted \texttt{cuad\_covenant\_not\_to\_sue} as curated; the
figures above are recomputed from the committed per-task discordance
tables over the documented six-task random sample, and reconcile with
Table 6.10.B, whose all-nine-task result is unaffected.)

The interpretation: the Eligibility Module's structural advantage is
\textbf{largest where the underlying rule has multiple
genuinely-distinct prongs} combined by deterministic logic --- exactly
the failure pattern of §4.2. On single-clause classification tasks the
structural advantage shrinks to a 1--3 percentage point edge that is
directionally consistent across tasks but small per task; at the
random-only level the combined test detects it clearly only against
GPT-5.4 --- the all-nine-task combination in Table 6.10.B is the primary
aggregate.

\subsubsection{6.10.4 GPT-5.4 prompt-format sensitivity on
classification
tasks}\label{gpt-5.4-prompt-format-sensitivity-on-classification-tasks}

GPT-5.4's accuracy on five of the six non-curated random-sample tasks
(\texttt{cuad\_covenant\_not\_to\_sue} 53.9\%,
\texttt{contract\_nli\_explicit\_id} 36.4\%,
\texttt{contract\_nli\_notice\_on\_compelled\_disclosure} 50.7\%,
\texttt{learned\_hands\_health} 49.6\%, \texttt{opp115\_international}
66.9\%) is sharply lower than Sonnet 4.6 and Opus 4.7 on the same tasks
(each clears 87\% on all five). Inspection of the per-case JSON
responses shows GPT-5.4 has a systematic positive-class bias on these
classification tasks: on \texttt{cuad\_covenant\_not\_to\_sue} it
returns ``yes'' on 304/308 test cases regardless of clause content
(recall \(\approx 100\%\), precision \(\approx 50\%\)). The pattern
replicates on the other four tasks and is not present on the multi-prong
rule tasks (\texttt{hearsay} 85.1\%, \texttt{personal\_jurisdiction}
96.0\%, \texttt{jcrew\_blocker} 92.6\%).

We do not characterise this as a capability statement about GPT-5.4
broadly, and we do not use these cells as primary evidence for the
exception-chain claim. The pattern is specific to the LegalBench
classification-style few-shot prompt format used here (the upstream
canonical \texttt{base\_prompt.txt}); alternative prompt strategies
(chain-of-thought, structured-output, zero-shot rule-only) may recover
the model's performance.

This distinction matters because the same model recovers under a simpler
prompt. We tested \texttt{cuad\_covenant\_not\_to\_sue} with a zero-shot
variant: the same model and the same task description from
\texttt{sources/rule.md}, but no few-shot Q/A examples --- only the
rule, the clause, and the question ``Answer Yes or No.'' On the same
154-case held-out subset, GPT-5.4 zero-shot achieves \textbf{143/154
(92.9\%)}, up from 83/154 (53.9\%) under the few-shot prompt --- a +39pp
recovery. The yes-rate normalises from 98.7\% under the few-shot prompt
to 45.5\%, matching the dataset's underlying \textasciitilde50\%
positive prior. This empirically confirms the framing for this task: the
low GPT-5.4 few-shot result is \textbf{prompt-format-coupled, not a
model-capability ceiling}.

Notably, the engine still beats GPT-5.4 zero-shot on this task: 150/154
(97.4\%) vs 143/154 (92.9\%), exact two-sided McNemar's \emph{b} = 7,
\emph{c} = 0, \emph{p} = 0.016. The margin holds even against the
best-prompt configuration of the strongest LLM tested.

\textbf{Figure 8: GPT-5.4 calibration cliff and zero-shot recovery.}

\pandocbounded{\includegraphics[keepaspectratio,alt={Five horizontal bars on the cuad\_covenant\_not\_to\_sue 154-case held-out subset. Engine (97.4\%), Sonnet 4.6 few-shot (96.1\%), and Opus 4.7 few-shot (96.8\%) cluster at the top; GPT-5.4 few-shot collapses to 53.9\% with a 98.7\% positive-class anchoring rate; under a zero-shot prompt (rule + clause + Yes/No, no Q/A examples) GPT-5.4 recovers to 92.9\% with a 45.5\% positive rate that matches the dataset prior. Wilson 95\% CIs as whiskers. Engine vs GPT-5.4 zero-shot paired McNemar's: b = 7, c = 0, p = 0.016. Generated by figures/scripts/cliff\_zero\_shot\_recovery.py.}]{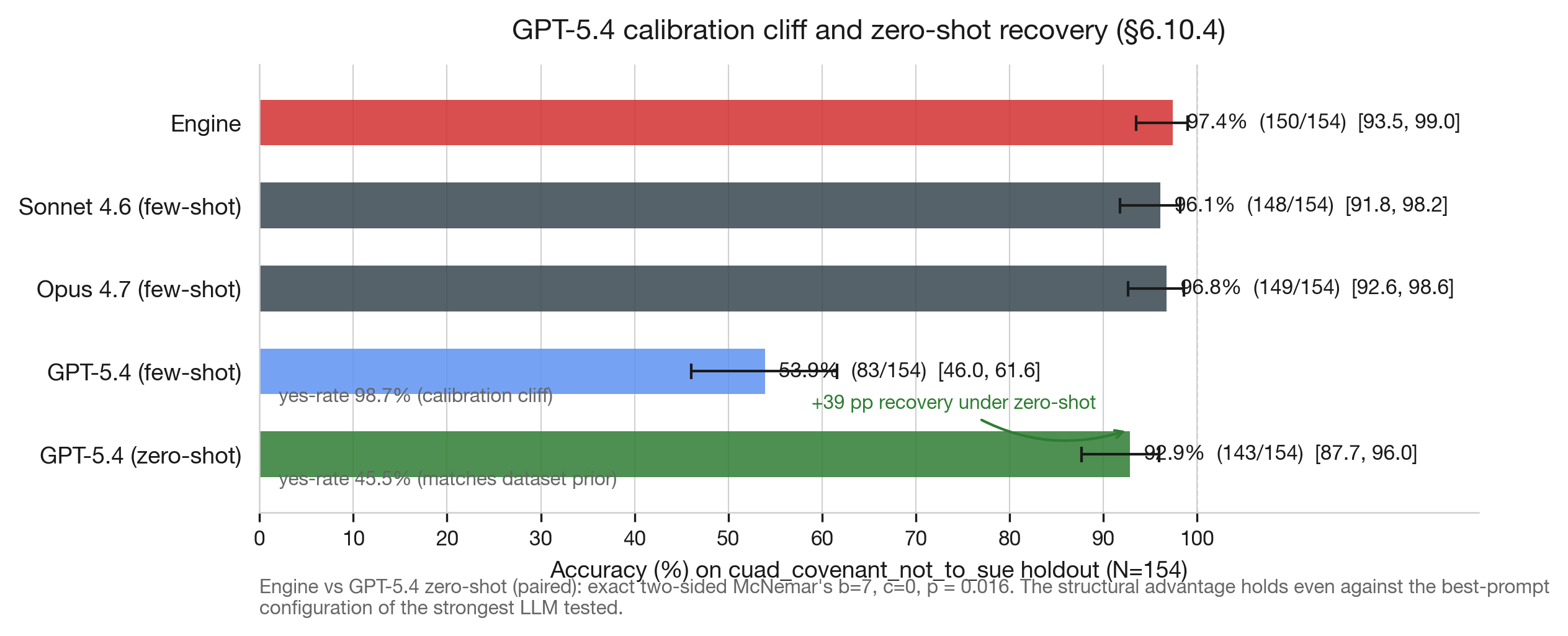}}

The relevant finding for the present paper is narrower. The multi-prong
tasks support the architectural claim about compositional rule
evaluation. The classification-prompt results document a separate
prompt-format sensitivity in the evaluated GPT-5.4 baseline. Both are
relevant to legal classification workflows, but only the former is
evidence for the exception-chain failure pattern.

\subsubsection{6.10.5 Cross-extractor: structural advantage is not
extractor-specific}\label{cross-extractor-structural-advantage-is-not-extractor-specific}

A second possible critique: the Eligibility Module pipeline uses an LLM
extractor (Sonnet 4.6 in the headline configuration) to produce the
typed field values that the engine evaluates. If the LLM-only baseline
also uses Sonnet 4.6, the comparison reduces to ``Sonnet split across
two steps versus Sonnet doing it all in one step'', which might be a
prompt-engineering artefact rather than a structural property of the
Module.

To test this, we re-ran the Eligibility Module pipeline with
\textbf{Opus 4.7 as the runtime extractor} on the two clearest
multi-prong tasks (\texttt{hearsay} and
\texttt{personal\_jurisdiction}), and compared against Opus 4.7
end-to-end:

\textbf{Table 6.10.C --- Cross-extractor (full test split, N as shown)}

{\def\LTcaptype{none} % do not increment counter
\begin{longtable}[]{@{}
  >{\raggedright\arraybackslash}p{(\linewidth - 8\tabcolsep) * \real{0.1579}}
  >{\raggedleft\arraybackslash}p{(\linewidth - 8\tabcolsep) * \real{0.2105}}
  >{\raggedleft\arraybackslash}p{(\linewidth - 8\tabcolsep) * \real{0.2105}}
  >{\raggedleft\arraybackslash}p{(\linewidth - 8\tabcolsep) * \real{0.2105}}
  >{\raggedleft\arraybackslash}p{(\linewidth - 8\tabcolsep) * \real{0.2105}}@{}}
\toprule\noalign{}
\begin{minipage}[b]{\linewidth}\raggedright
Task
\end{minipage} & \begin{minipage}[b]{\linewidth}\raggedleft
Sonnet 4.6 end-to-end
\end{minipage} & \begin{minipage}[b]{\linewidth}\raggedleft
Sonnet-extractor + engine
\end{minipage} & \begin{minipage}[b]{\linewidth}\raggedleft
Opus 4.7 end-to-end
\end{minipage} & \begin{minipage}[b]{\linewidth}\raggedleft
Opus-extractor + engine
\end{minipage} \\
\midrule\noalign{}
\endhead
\bottomrule\noalign{}
\endlastfoot
\texttt{hearsay} (94) & 76.6\% & 90.4\% & 83.0\% & 86.2\% \\
\texttt{personal\_jurisdiction} (50) & 92.0\% & 98.0\% & 92.0\% &
98.0\% \\
\end{longtable}
}

The advantage holds for \textbf{both} extractor models. On
\texttt{personal\_jurisdiction}, the lift is identical (+6.0pp)
regardless of extractor; on \texttt{hearsay}, the lift is +13.8pp with
Sonnet and +3.2pp with Opus --- Opus is genuinely better at hearsay
reasoning end-to-end, so the symbolic engine adds less incremental value
to the stronger extractor. In neither case does the Module pipeline lose
to its corresponding LLM-only baseline. The result is consistent with
the §4.2 framing: the deterministic conjunction step is the source of
the gain, not the choice of extractor.

\subsubsection{6.10.6 Discussion}\label{discussion}

The LegalBench results are external validation of the central claim of
this paper: on multi-prong rule-evaluation tasks, the Eligibility
Module's deterministic-execution architecture is more accurate than
end-to-end LLM evaluation, and the gap is reliably detectable. The gap
shrinks to a small but consistent margin on single-clause classification
tasks chosen at random --- the regime where there is no compositional
rule structure for the symbolic engine to exploit --- and the gap holds
across two LLM model families (Anthropic, OpenAI) and across two choices
of runtime extractor (Sonnet 4.6, Opus 4.7).

The retroactive held-out methodology is acknowledged as a methodological
compromise. For the three curated tasks, hint iteration during authoring
used the full LegalBench \texttt{test} split before the held-out
partition was introduced; the held-out results in Table 6.10.A are
therefore reported on a half-test subset that the authors \emph{did} see
during iteration but did not focus on. For the six random-sample tasks,
the held-out methodology was applied prospectively (hints iterated only
on a dev half, holdout evaluated once with frozen hints). The combined
paired-binomial result in Table 6.10.B is dominated by the random-sample
tasks by case count (850 of 949 cases) and remains significant.

\textbf{Engine error analysis.} The engine is correct on 900/949
held-out cases (94.8\%); the 49 errors are not uniformly distributed. By
task: \texttt{opp115\_international\_and\_specific\_audiences} 23/347
(6.6\%, the largest absolute contributor),
\texttt{contract\_nli\_explicit\_identification} 7/55 (12.7\%, the
highest rate), \texttt{learned\_hands\_health} 6/113 (5.3\%),
\texttt{cuad\_liquidated\_damages} 5/110 (4.5\%),
\texttt{cuad\_covenant\_not\_to\_sue} 4/154 (2.6\%), \texttt{hearsay}
2/47 (4.3\%), \texttt{personal\_jurisdiction} 1/25 (4.0\%),
\texttt{contract\_nli\_notice\_on\_compelled\_disclosure} 1/71 (1.4\%),
\texttt{jcrew\_blocker} 0/27 (0.0\%). The three curated multi-prong
tasks contribute 3 errors out of 99 cases (3.0\%); the six single-clause
classification (random-sample) tasks contribute 46 errors out of 850
cases (5.4\%). The pattern is consistent with the structural-advantage
interpretation in §6.10.3: where the underlying rule has multiple
genuinely-distinct prongs and the engine's deterministic conjunction
step is structurally aligned with the rule shape, errors are rare. Where
the task is single-clause classification and the engine reduces to ``the
LLM extractor's binary judgement, run through a one-clause rule,'' the
engine's accuracy reflects the extractor's accuracy on borderline cases.

\textbf{Model-family and jurisdictional limitations.} This evaluation
tests two model families (Anthropic via Sonnet 4.6 and Opus 4.7; OpenAI
via GPT-5.4). It does not test other frontier or production LLMs (Llama,
Gemini, DeepSeek, Mistral). The §6.10.4 calibration finding for GPT-5.4
is shown to be prompt-format-coupled within OpenAI's GPT-5 family but
its presence or absence in other model families is untested. All nine
LegalBench tasks evaluated here are US-jurisdiction English-language
tasks; the §6 controlled benchmark is UK-leaning. Generalisation to
other jurisdictions or non-English legal corpora is out of scope for the
present revision.

\textbf{What would refute this section's central claim.} A documented
LLM-only configuration that, on the same 949 held-out LegalBench cases
(with the deterministic seed-7 partition reproduced from
\texttt{tools/test\_split.py}), achieves a combined accuracy ≥ 95\% on
the same nine tasks under any disclosed prompting strategy would
substantially weaken the structural-advantage interpretation in
§6.10.5--§6.10.6. Such a result would not falsify the existence of the
§4.2 failure pattern (which is a small-N existence claim by design) but
would show that prompt engineering alone can match the Module on this
evaluation set, removing the \emph{architectural} basis for the headline
gap reported in Table 6.10.B.

The benchmark scenarios, source files, runners, statistical analysis
script, and pre-registration artefacts are released in the
\texttt{legalbench/} subdirectory of
\url{https://github.com/Aethis-ai/confidently-wrong-benchmark} for
replication. The pre-registration of the seed-44 random-sample
replication is verifiable via the public Git tag
\texttt{pre-v3.8-legalbench-preregistration} (annotated tag object
\texttt{f7c5994}, tagged commit \texttt{58d3b5d}). The held-out
partition reproducibility is verified at
\texttt{legalbench/docs/reproducibility-checks.md} (9/9 tasks:
re-running
\texttt{tools/test\_split.py\ -\/-seed\ 7\ -\/-dev\_fraction\ 0.5} from
a clean checkout reproduces the index sets used in committed engine
results exactly).

\begin{center}\rule{0.5\linewidth}{0.5pt}\end{center}

\section{7. Challenges in LLM-Guided Rule
Synthesis}\label{challenges-in-llm-guided-rule-synthesis}

Rule authoring is the bottleneck. The execution layer (Level 3) provides
deterministic guarantees, but those guarantees are only as good as the
rules they operate on. Execution correctness is a necessary precondition
for trust --- it eliminates one entire class of error --- but the hard
problem remains: can an LLM reliably formalise legislation into rules
that faithfully capture the legislator's intent? This paper does not
claim to solve that problem. It reports that the execution layer is
robust and that we approach the Level 2 problem through two
complementary quality mechanisms, summarised below. Implementation
specifics --- thresholds, iteration bounds, internal evaluator rubrics,
and feedback-steering heuristics --- are not disclosed here.

\subsection{7.1 Iterative LLM-Assisted
Authoring}\label{iterative-llm-assisted-authoring}

Rules are authored through iterative synthesis with automated validation
and convergence criteria: a frontier LLM generates candidate rules from
authoritative source text, receives structured critique, and refines on
successive passes until quality has stabilised. This is a
research-engineering component, not a correctness guarantee: the purpose
of Section 5.5 and Section 7.2 is to make clear which guarantees we do
claim and which we do not.

\subsection{7.2 Honest Assessment}\label{honest-assessment}

These quality scores are produced by an LLM evaluator, not by
independent human legal experts. They measure internal consistency
against the evaluator's rubric, not ground-truth correctness. The
test-driven validation layer (Section 7.3) provides the stronger signal,
but test suites are only as comprehensive as the scenarios SMEs define.
This section documents engineering progress on a difficult problem; it
is not a claim of solved correctness at Level 2.

\subsection{7.3 Test-Driven Validation}\label{test-driven-validation}

Rule quality is gated against concrete expected outcomes, not against
LLM self-evaluation alone. Subject matter experts author test suites
that function as acceptance criteria for the authored rule bundle.

\textbf{Test case structure.} Each case pairs structured applicant data
with an expected outcome (eligible / ineligible / undetermined) for a
specific legal section. Cases are tagged by category --- golden path,
edge case, and corner case --- with corner cases specifically exercising
exception chains and override logic.

\textbf{Validation and regression.} When rules are authored or revised,
the bundle is compiled and every test case is evaluated through the
constraint engine. Failures produce structured diagnostics --- which
constraint, which expected outcome, which branch of the disjunction ---
that feed back into the authoring loop. A golden-case suite is preserved
across rule versions so that re-authoring (due to legislation changes or
quality improvements) cannot silently regress a previously-correct
scenario. This outer loop, grounded in SME-defined outcomes rather than
LLM self-assessment, is the primary quality gate for deployment.

\begin{center}\rule{0.5\linewidth}{0.5pt}\end{center}

\section{8. Generalisation and
Applicability}\label{generalisation-and-applicability}

\subsection{8.1 Domain-Agnostic
Architecture}\label{domain-agnostic-architecture}

The system is architecturally domain-agnostic. An internal audit found
that over 90\% of the codebase is shared across domains. The only
components requiring replacement for a new vertical are the source
connectors (which fetch and parse authoritative text from
domain-specific APIs) and any domain-specific business logic. The core
engine, rule representation, compilation pipeline, quality evaluation,
and API layer work unchanged.

\subsection{8.2 Domain Injection Points}\label{domain-injection-points}

{\def\LTcaptype{none} % do not increment counter
\begin{longtable}[]{@{}
  >{\raggedright\arraybackslash}p{(\linewidth - 4\tabcolsep) * \real{0.2982}}
  >{\raggedright\arraybackslash}p{(\linewidth - 4\tabcolsep) * \real{0.2982}}
  >{\centering\arraybackslash}p{(\linewidth - 4\tabcolsep) * \real{0.4035}}@{}}
\toprule\noalign{}
\begin{minipage}[b]{\linewidth}\raggedright
Injection Point
\end{minipage} & \begin{minipage}[b]{\linewidth}\raggedright
What It Provides
\end{minipage} & \begin{minipage}[b]{\linewidth}\centering
Manual or Discovered?
\end{minipage} \\
\midrule\noalign{}
\endhead
\bottomrule\noalign{}
\endlastfoot
Source connectors & Authoritative text for the domain & Manual
(one-time) \\
Guidance hints & Subject matter expert steering & Manual (optional) \\
Field registry & Typed data fields that rules reference & LLM-assisted
discovery \\
Rules and criteria & Logical conditions compiled to constraints &
LLM-generated, human-reviewable \\
Quality rubric & Evaluation criteria for generated rules &
Domain-agnostic (built-in) \\
\end{longtable}
}

The field registry and rules are outputs of the pipeline, not inputs.
The source connectors are the only early-stage domain-specific code, and
they are modular: adapting to a new domain means replacing one source
connector with an equivalent for another regulatory body.

\subsection{8.3 Adjacent Verticals}\label{adjacent-verticals}

The architecture is applicable to any domain exhibiting the three
properties identified in Section 4.1: material consequence,
auditability, and OR-branching logic:

\begin{itemize}
\tightlist
\item
  \textbf{Benefits and welfare eligibility:} Universal Credit,
  disability benefits, pension entitlements with complex exemption and
  transition rules
\item
  \textbf{Tax compliance:} R\&D tax credits, corporate tax relief
  pathways with safe harbours
\item
  \textbf{Financial regulatory compliance:} MiFID II suitability, AML
  threshold assessments
\item
  \textbf{Insurance claims adjudication:} Coverage determination under
  exclusion clauses with nested exception chains (benchmarked: Section
  6.4)
\item
  \textbf{Safety certification:} Aviation crew certification, equipment
  compliance with conditional overrides
\item
  \textbf{Healthcare pathway determination:} Clinical pathway
  eligibility, treatment authorisation chains
\end{itemize}

\begin{center}\rule{0.5\linewidth}{0.5pt}\end{center}

\section{9. Compliance and Regulatory
Considerations}\label{compliance-and-regulatory-considerations}

The EU AI Act \citeyearpar{euaiact2024} classifies as high-risk certain
specified AI uses in migration, asylum, visa, and residence contexts
when deployed by or on behalf of competent public authorities (Annex
III, point 7), requiring conformity assessments, risk management
systems, and human oversight. Whether a given deployment falls within
Annex III depends on the specific use and deploying entity; we do not
claim that the system described here is automatically in scope. More
broadly, systems used for eligibility determination in regulated domains
face increasing scrutiny under this and similar frameworks, and the
architecture described here - deterministic execution with full
provenance - is designed with those requirements in mind.

\subsection{9.1 Auditability}\label{auditability}

Every determination produced by the system is fully auditable through a
provenance chain:

\begin{enumerate}
\def\labelenumi{\arabic{enumi}.}
\tightlist
\item
  \textbf{Determination:} the Eligibility Module returns
  eligible/ineligible with the specific constraints satisfied or
  violated. The determination result itself carries the provenance
  chain: each criterion that contributed to the outcome includes its
  source citations (document ID, section path, and direct quote from the
  authoritative text). This means the provenance is not a separate
  lookup --- it is returned alongside the determination.
\item
  \textbf{Rule:} each constraint maps to a named rule in the rule
  bundle, with its rule structure visible.
\item
  \textbf{Source citation:} each rule carries citations linking it to
  specific legislative clauses with direct quotes and section path
  references. During rule synthesis, the LLM's context includes tagged
  source items (e.g., \texttt{{[}BNA1981\#Schedule1/P1.1{]}}), and the
  LLM cites these tags directly on each generated criterion. A
  provenance verifier resolves these citations against the assembled
  source material, creating structured anchors with document IDs,
  section paths, and direct quotes. When the LLM omits citations or
  references invalid sources, the verifier falls back to embedding-based
  semantic similarity as a secondary mechanism. This multi-stage
  citation verification with semantic fallback is more reliable than
  post-hoc matching alone, though the fallback path remains
  probabilistic and may occasionally misattribute a rule to a closely
  related but incorrect clause.
\item
  \textbf{Legislation:} the original legislative text is available via
  legislation.gov.uk and GOV.UK APIs.
\end{enumerate}

\textbf{Figure 6: Provenance Chain (Two-Stage Verification)}

\pandocbounded{\includegraphics[keepaspectratio,alt={Provenance Chain (Two-Stage Verification)}]{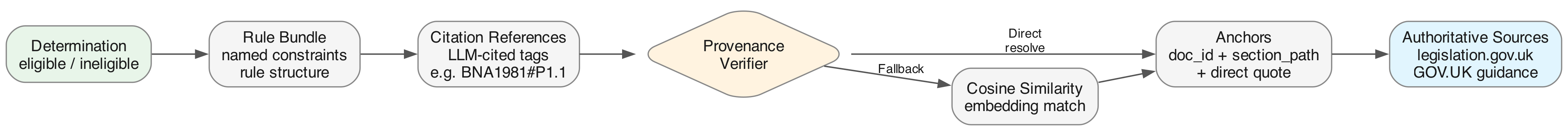}}

This chain is qualitatively different from an LLM's reasoning trace. An
LLM produces a natural language explanation that may not reflect its
actual inference process. The engine produces the exact logical path:
the specific constraints evaluated, the specific values tested, and the
specific branch of the OR expression that was or was not satisfied.

\subsection{9.2 Explainability and Human
Oversight}\label{explainability-and-human-oversight}

The Module produces a binary determination with the exact logical path,
not a confidence score. A solicitor can verify the determination by
checking the logical path against their understanding of the
legislation. The system supports human-supervised workflows and can be
configured so that determinations are reviewed by a professional before
external action is taken.

\subsection{9.3 Quality Gates}\label{quality-gates}

The rule authoring pipeline includes multiple quality gates before a
rule bundle is persisted for deployment: iterative LLM
synthesis-and-critique (Section 7.1), source citation validation, and
--- as the strongest gate --- test-driven validation against SME-defined
test suites (Section 7.3). Rules must produce correct outcomes on golden
path, edge case, and corner case scenarios defined by domain experts
before they are deployed.

\subsection{9.4 Model Drift and Bundle
Certification}\label{model-drift-and-bundle-certification}

Frontier LLMs change between API versions and, as the v3.8 replication
documents (§6.5 Finding 4), within the same nominal model alias without
announcement. For systems that perform inference on every query against
a frontier LLM, this raises a regulatory question with no clean answer:
how can a deployed determination system be certified at time T when the
underlying inference engine can change at time T+1 without notice or
visibility? The architecture separates this question into two parts that
can be addressed independently.

\textbf{Execution layer (L3) is invariant.} Once a rule bundle is
compiled, every determination is produced by deterministic SMT-based
evaluation against that bundle. No LLM is in the inference path.
Frontier model changes --- silent or announced --- do not alter the
determinations produced by an existing bundle on existing inputs. A
regulator certifying a deployed bundle is certifying a fixed artefact,
not a moving target.

\textbf{Authoring layers (L1 retrieval, L2 formalisation) are versioned
and replayable.} Each rule bundle is generated against a recorded model
snapshot, with full provenance back to source legislation. If a future
model would generate different rules from the same sources, the deployed
bundle is unaffected; re-authoring is a deliberate, audited action that
re-triggers test-driven validation (Section 7.3) and produces a new
bundle version. SME test suites act as a regression gate: silent
behavioural changes in the authoring model that affect rule semantics
surface as test failures rather than as silent eligibility drift in
production.

This separation does not eliminate the underlying drift problem in
frontier models --- it relocates it to the authoring boundary, where it
is observable and auditable, rather than the inference boundary, where
it is silent and continuous. Whether the L1/L2 boundary is itself
sufficiently controlled for a given regulated workflow is a question the
bundle's quality engineering must answer (Sections 7 and 9.3); a
separate open question is whether silent drift in the authoring model
can produce silent incorrect rules that pass the SME test suite by
coincidence --- this is tracked as an open follow-up item rather than
resolved within this revision.

\begin{center}\rule{0.5\linewidth}{0.5pt}\end{center}

\section{10. Limitations and Future
Work}\label{limitations-and-future-work}

\subsection{10.1 Current Limitations}\label{current-limitations}

\begin{enumerate}
\def\labelenumi{\arabic{enumi}.}
\tightlist
\item
  \textbf{Structured input only.} The benchmark tests legal reasoning
  accuracy, not extraction from natural language or documents. A
  production system requires both; the extraction layer introduces its
  own error surface.
\item
  \textbf{Level 2 validation is test-suite-dependent.} Rule
  formalisation quality is validated against SME-defined test suites
  (Section 7.3) and measured by internal LLM evaluator scores. The test
  suites provide ground-truth validation but are only as comprehensive
  as the scenarios they cover. Independent human review of generated
  rules against authoritative legal interpretations has not been
  performed.
\item
  \textbf{225 scenarios across four domains.} The benchmark does not
  exhaust legal reasoning. It tests one class of task across four
  domains; findings should not be generalised beyond nested
  exception-chain evaluation.
\item
  \textbf{Model versioning.} Results are tied to specific model versions
  available at time of evaluation. The public benchmark dataset allows
  this to be re-evaluated as models improve.
\item
  \textbf{Benchmark performance is not a production guarantee.} The
  accuracy figures reported here reflect performance on fixed benchmark
  scenarios under controlled conditions. They do not establish
  correctness for any specific real-world legal determination.
  Production deployment of an automated eligibility or compliance system
  requires professional legal review per jurisdiction, ongoing
  monitoring of model and legislative changes, and appropriate human
  oversight. Nothing in this paper should be read as legal advice or as
  a warranty of fitness for a particular legal purpose.
\item
  \textbf{The failure mode is bounded.} The systematic failures
  characterised here concentrate at three-level exception depth and on
  the multi-prong compositional sub-task (the precise scope phrase
  introduced in §1: \emph{multi-prong compositional rule evaluation
  under nested exceptions}). On shallower or non-exception eligibility
  logic, frontier-model accuracy on this benchmark is high, and §6.4.1
  documents a declining adversarial failure rate as frontier models
  update. The contribution is the demonstration of a durable boundary
  condition where deterministic execution is required, not a claim of
  broad LLM unreliability.
\end{enumerate}

\subsection{10.2 Future Work}\label{future-work}

{\def\LTcaptype{none} % do not increment counter
\begin{longtable}[]{@{}
  >{\centering\arraybackslash}p{(\linewidth - 2\tabcolsep) * \real{0.6250}}
  >{\raggedright\arraybackslash}p{(\linewidth - 2\tabcolsep) * \real{0.3750}}@{}}
\toprule\noalign{}
\begin{minipage}[b]{\linewidth}\centering
Priority
\end{minipage} & \begin{minipage}[b]{\linewidth}\raggedright
Item
\end{minipage} \\
\midrule\noalign{}
\endhead
\bottomrule\noalign{}
\endlastfoot
High & Human evaluation calibration: compare LLM quality scores against
human-labelled rule sets \\
High & Multi-section regression suite: automated quality runs across all
sections \\
Medium & Natural language input testing: extend benchmark to
unstructured applicant narratives \\
Medium & Multi-model evaluation ensemble: cross-validate quality scores
across model families \\
Low & Independent legal review: solicitor verification of expected
values \\
\end{longtable}
}

\begin{center}\rule{0.5\linewidth}{0.5pt}\end{center}

\section{Acknowledgments}\label{acknowledgments}

We thank an anonymous reviewer (referee report received 2026-04-28) for
the framing of the architectural contribution as a \emph{relocation} of
uncertainty rather than its elimination. That framing clarified the
rhetorical centre of gravity of §2, §5.5, and §9.4, and produced the
boundary-condition restatement of the failure mode in §10.1. We also
thank the project's adversarial-review process --- both human and
automated --- for sustained pressure on the specification-vs-execution
distinction, the prompt-engineering claim, and the scope of the failure
regime characterised here.

\begin{center}\rule{0.5\linewidth}{0.5pt}\end{center}

\section{References}\label{references}

\begin{center}\rule{0.5\linewidth}{0.5pt}\end{center}

\section{Technical Appendix A: Benchmark
Methodology}\label{technical-appendix-a-benchmark-methodology}

\subsection{A.1 Life in the UK Scenario
Matrix}\label{a.1-life-in-the-uk-scenario-matrix}

{\def\LTcaptype{none} % do not increment counter
\begin{longtable}[]{@{}lll@{}}
\toprule\noalign{}
Field & Type & Values Tested \\
\midrule\noalign{}
\endhead
\bottomrule\noalign{}
\endlastfoot
\texttt{test\_passed} & Boolean & True, False \\
\texttt{medical\_exempt} & Boolean & True, False \\
\texttt{discretion\_applied} & Boolean & True, False \\
\texttt{applicant\_age} & Integer & 15, 17, 18, 30, 64, 65, 70 \\
\end{longtable}
}

\textbf{Total combinations:} 2 x 2 x 2 x 7 = 56 scenarios

\textbf{Expected value formula:}
\texttt{eligible\ =\ test\_passed\ OR\ (age\ \textless{}\ 18)\ OR\ (age\ \textgreater{}=\ 65)\ OR\ medical\_exempt\ OR\ discretion\_applied}

\subsection{A.2 English Language Scenario
Design}\label{a.2-english-language-scenario-design}

43 scenarios (24 baseline + 19 adversarial): MESC nationality route;
Canada bilingual special case; SELT expiry rules; degree certification;
near-miss combinations; multi-field distractors; age boundary stacking.

\subsection{A.3 Spacecraft Scenario
Design}\label{a.3-spacecraft-scenario-design}

68 scenarios (48 baseline + 20 adversarial) using a synthetic statute
modelled on UK legislative structure. Exercises eight patterns: UNSAT
early termination, multi-field AND, multi-route OR, three-level
exception chain, nested IMPLIES, ENUM + IN, compound multi-requirement,
adversarial traps (veteran age-independence, IMPLIES vacuous truth,
direct satisfaction vs.~exemption).

\subsection{A.4 Construction Insurance Scenario
Design}\label{a.4-construction-insurance-scenario-design}

58 scenarios using a synthetic CAR policy defect exclusion endorsement
modelled on London market DE3/DE5 clause structure. Exercises eight
patterns: UNSAT early termination, multi-field AND, multi-route OR,
five-level exception chain, nested IMPLIES, ENUM + IN, DATE-bounded
BOOL, absolute exclusion, plus demo, adversarial, and boundary-stacking
scenarios.

\textbf{Fields:}

{\def\LTcaptype{none} % do not increment counter
\begin{longtable}[]{@{}
  >{\raggedright\arraybackslash}p{(\linewidth - 4\tabcolsep) * \real{0.3333}}
  >{\raggedright\arraybackslash}p{(\linewidth - 4\tabcolsep) * \real{0.2857}}
  >{\raggedright\arraybackslash}p{(\linewidth - 4\tabcolsep) * \real{0.3810}}@{}}
\toprule\noalign{}
\begin{minipage}[b]{\linewidth}\raggedright
Field
\end{minipage} & \begin{minipage}[b]{\linewidth}\raggedright
Type
\end{minipage} & \begin{minipage}[b]{\linewidth}\raggedright
Values
\end{minipage} \\
\midrule\noalign{}
\endhead
\bottomrule\noalign{}
\endlastfoot
\texttt{car.policy.period\_valid} & Boolean & True, False \\
\texttt{car.property.category} & Enum & permanent\_works,
temporary\_works, plant\_equipment, existing\_structures,
materials\_on\_site \\
\texttt{car.loss.is\_physical} & Boolean & True, False \\
\texttt{car.component.is\_defective} & Boolean & True, False \\
\texttt{car.defect.origin} & Enum & design, specification, materials,
workmanship, none \\
\texttt{car.claim.is\_rectification} & Boolean & True, False \\
\texttt{car.claim.is\_access\_damage} & Boolean & True, False \\
\texttt{car.damage.consequence\_of\_failure} & Boolean & True, False \\
\texttt{car.project.value\_millions\_gbp} & Integer & 1--800 \\
\texttt{car.notification.within\_period} & Boolean & True, False \\
\texttt{car.contract.jct\_compliant} & Boolean & True, False \\
\texttt{car.defect.was\_known\_prior} & Boolean & True, False \\
\texttt{car.defect.has\_engineer\_assessment} & Boolean & True, False \\
\texttt{car.component.installed\_by\_subcontractor} & Boolean & True,
False \\
\end{longtable}
}

\subsection{A.5 LLM Prompt Structure}\label{a.5-llm-prompt-structure}

\begin{enumerate}
\def\labelenumi{\arabic{enumi}.}
\tightlist
\item
  Full section text (markdown-formatted legislation and guidance)
\item
  Structured applicant data as key-value pairs
\item
  System prompt: \emph{``You are a legal eligibility assessor. Given the
  legislation and applicant data, determine whether the applicant meets
  the requirement. Respond with JSON:
  \texttt{\{"eligible":\ true/false,\ "reasoning":\ "..."\}}''}
\end{enumerate}

No special instructions about exemptions, alternative routes, or
OR-branching.

\subsection{A.6 Reproduction}\label{a.6-reproduction}

\begin{Shaded}
\begin{Highlighting}[]
\CommentTok{\# Full benchmark (225 scenarios across all sections)}
\ExtensionTok{uv}\NormalTok{ run python scripts/benchmark\_accuracy\_ab.py }\AttributeTok{{-}{-}model}\NormalTok{ claude{-}opus{-}4{-}6}

\CommentTok{\# Single section}
\ExtensionTok{uv}\NormalTok{ run python scripts/benchmark\_accuracy\_ab.py }\AttributeTok{{-}{-}model}\NormalTok{ gpt{-}5.4 }\AttributeTok{{-}{-}section}\NormalTok{ spacecraft}

\CommentTok{\# Construction insurance comparison}
\ExtensionTok{uv}\NormalTok{ run python scripts/llm\_vs\_engine\_comparison.py }\AttributeTok{{-}{-}models} \StringTok{"gpt{-}5.4,gpt{-}4.1{-}mini"}

\CommentTok{\# More LLM runs for higher confidence}
\ExtensionTok{uv}\NormalTok{ run python scripts/benchmark\_accuracy\_ab.py }\AttributeTok{{-}{-}model}\NormalTok{ gpt{-}5{-}mini }\AttributeTok{{-}{-}llm{-}runs}\NormalTok{ 5}

\CommentTok{\# Parallel evaluation (faster for large scenario counts)}
\ExtensionTok{uv}\NormalTok{ run python scripts/benchmark\_accuracy\_ab.py }\AttributeTok{{-}{-}model}\NormalTok{ gpt{-}5.4 }\AttributeTok{{-}{-}concurrency}\NormalTok{ 10}
\end{Highlighting}
\end{Shaded}

Supported models: \texttt{gpt-5.4}, \texttt{gpt-5.3}, \texttt{gpt-5.2},
\texttt{gpt-5-mini}, \texttt{gpt-5-nano}, \texttt{gpt-4.1-mini},
\texttt{claude-opus-4-7}, \texttt{claude-opus-4-6},
\texttt{claude-sonnet-4-6}, \texttt{claude-haiku-4-5},
\texttt{gemini-3.1-pro}, \texttt{gemini-3.1-flash-lite},
\texttt{deepseek-r1}.

\begin{center}\rule{0.5\linewidth}{0.5pt}\end{center}

\section{Technical Appendix B: Architecture
Overview}\label{technical-appendix-b-architecture-overview}

The Eligibility Module uses a typed rule representation that is parsed
from a constrained DSL, compiled to formal constraints, and evaluated by
an SMT-based constraint evaluation engine. Rules are organised into
groups representing alternative legal routes (OR'd within groups, AND'd
across groups), with support for custom outcome logic including material
implication.

The compilation pipeline transforms authored rules into a live
constraint evaluation engine capable of deterministic evaluation,
dynamic question generation, and provenance tracking from determination
back to source legislation.

All rules, fields, and rule sets are stored immutably with versioning
and supersedes references. This architecture detail is available under
NDA for technical due diligence.

\begin{center}\rule{0.5\linewidth}{0.5pt}\end{center}

\section{Author Contributions}\label{author-contributions}

P.S. conceived the paper, designed and implemented the Eligibility
Module --- including the code-generation layer, compilation
architecture, SMT constraint-evaluation engine, rule-authoring tools,
and benchmark harness --- designed and ran all experiments, analysed the
results, and drafted the manuscript. P.S. additionally designed and ran
the §6.10 LegalBench external-validation programme: per-task rule-bundle
authoring, SME-guidance design, runtime extractor configurations,
statistical-analysis pipeline (\texttt{tools/significance.py}), held-out
partition methodology, pre-registered random sampling protocol, GPT-5.4
zero-shot prompt-sensitivity check, and Figures 7 and 8. J.K. proposed
the use of an SMT solver as the execution engine for deterministic rule
evaluation within P.S.'s existing code-generation and compilation
architecture, and contributed ongoing design discussion. L.D.,
Immigration Solicitor and accredited Senior Caseworker and Supervisor
under the Law Society Immigration and Asylum Accreditation Scheme, with
over twenty years' experience in immigration law, authored the benchmark
scenarios for the British Nationality Act eligibility determinations
(\texttt{life\_uk} and \texttt{english\_language} sections) and
validated their legal formalisations; she did not contribute to the
spacecraft, construction insurance, or LegalBench materials. All authors
reviewed the final manuscript.

\section{Competing Interests}\label{competing-interests}

All authors are affiliated with Aethis, which develops the Eligibility
Module evaluated in this paper and is deploying it in a controlled UK
immigration pilot, in which it prepares eligibility evaluations for
solicitor review and solicitors remain the decision-makers. The
benchmark scenarios, harness, prompts, and per-case result artefacts are
publicly released (see Data and Code Availability) so that all reported
comparisons can be independently reproduced.

\section{Data and Code Availability}\label{data-and-code-availability}

The benchmark scenarios, expected outcomes, prompts, and evaluation
harness for the §6.1--§6.9 controlled benchmark (225 paper-scope
scenarios across four domains) are released publicly as the
\emph{confidently-wrong-benchmark} repository:
\url{https://github.com/Aethis-ai/confidently-wrong-benchmark}.

The §6.10 LegalBench external-validation harness (15 per-task projects
covering the 9 reported tasks plus 6 supplementary, including verbatim
CC-BY-4.0 source-rule files, all per-case engine + LLM result JSONs for
every reported number, the dev/holdout partition utility, the seeded
random-sample task picker, the per-task statistical-analysis pipeline,
and the seed-44 pre-registration tag) is released as the
\texttt{legalbench/} subdirectory of the same repository:
\url{https://github.com/Aethis-ai/confidently-wrong-benchmark/tree/main/legalbench}.
The migration is tagged as \texttt{pre-v3.8-legalbench-preregistration}
(annotated tag object \texttt{f7c5994}, tagged commit \texttt{58d3b5d})
--- that tag is the externally-verifiable boundary for the §6.10.3
random-sample-replication claim. Held-out partition reproducibility is
verified at \texttt{legalbench/docs/reproducibility-checks.md} (9/9
tasks reproduce exactly from a clean checkout).

Both sub-corpora are intended to allow independent re-evaluation of the
reported results and evaluation of future models against the same task
structure.

\begin{center}\rule{0.5\linewidth}{0.5pt}\end{center}

\begin{center}\rule{0.5\linewidth}{0.5pt}\end{center}

\emph{Version 3.13.0 · Working paper · July 2026}

\bibliography{references.bib}

\end{document}